\pdfoutput=1
\documentclass[11pt]{article}

\usepackage[]{acl}

\usepackage{times}
\usepackage{latexsym}
\usepackage{changepage}
\usepackage{multirow}
\usepackage{booktabs}
\usepackage{xspace}
\usepackage{amsfonts}
\usepackage{enumitem}
\usepackage{amsmath}
\usepackage{amssymb}
\usepackage{microtype}
\usepackage{pifont}
\usepackage{tikz}
\usepackage{colortbl}
\usepackage{url}
\usepackage{dsfont}
\usepackage{tabularx}
\usepackage[T1]{fontenc}
\usepackage[utf8]{inputenc}
\usepackage[normalem]{ulem}

\definecolor{green2}{HTML}{BFD8B6}
\definecolor{green3}{HTML}{E7F0E5}
\definecolor{greenarrow}{HTML}{1DB100}
\definecolor{red3}{HTML}{C82506}

\usepackage[T1]{fontenc}
\usepackage[utf8]{inputenc}

\usepackage{microtype}

\usepackage{inconsolata}

\newcommand{\benchmarkname}[1]{\textsc{LLM-AggreFact}}

\usepackage[textsize=scriptsize]{todonotes}

\title{MiniCheck: Efficient Fact-Checking of LLMs on Grounding Documents}

\author{Liyan Tang$^\diamondsuit$ \quad \quad \quad Philippe Laban$^\spadesuit$ \quad \quad \quad Greg Durrett$^\diamondsuit$ \\
        $^\diamondsuit$The University of Texas at Austin  \quad $^\spadesuit$Salesforce AI Research\\   
\texttt{lytang@utexas.edu}}

\begin{document}
\maketitle
\begin{abstract}
Recognizing if LLM output can be grounded in evidence is central to many tasks in NLP: retrieval-augmented generation, summarization, document-grounded dialogue, and more. Current approaches to this kind of fact-checking are based on verifying each piece of a model generation against potential evidence using an LLM. However, this process can be very computationally expensive, requiring many calls to a model to check a single response. In this work, we show how to build small fact-checking models that have GPT-4-level performance but for 400x lower cost. We do this by constructing synthetic training data with GPT-4, which involves creating realistic yet challenging instances of factual errors via a structured generation procedure. Training on this data teaches models to check each fact in the claim and recognize synthesis of information across sentences. For evaluation, we unify datasets from recent work on fact-checking and grounding LLM generations into a new benchmark, \benchmarkname{}. Our best system MiniCheck-\textsc{FT5} (770M parameters) outperforms all systems of comparable size and reaches GPT-4 accuracy. We release \benchmarkname{}, code for data synthesis, and models.\footnote{ \href{https://github.com/Liyan06/MiniCheck}{\texttt{https://github.com/Liyan06/MiniCheck}}; see latest leaderboard at \href{https://llm-aggrefact.github.io}{\texttt{llm-aggrefact.github.io}}.}
\end{abstract}

\section{Introduction}

\begin{figure}[t!]
\centering
\includegraphics[width=1\linewidth, trim=0mm 0mm 00mm 0mm,clip]{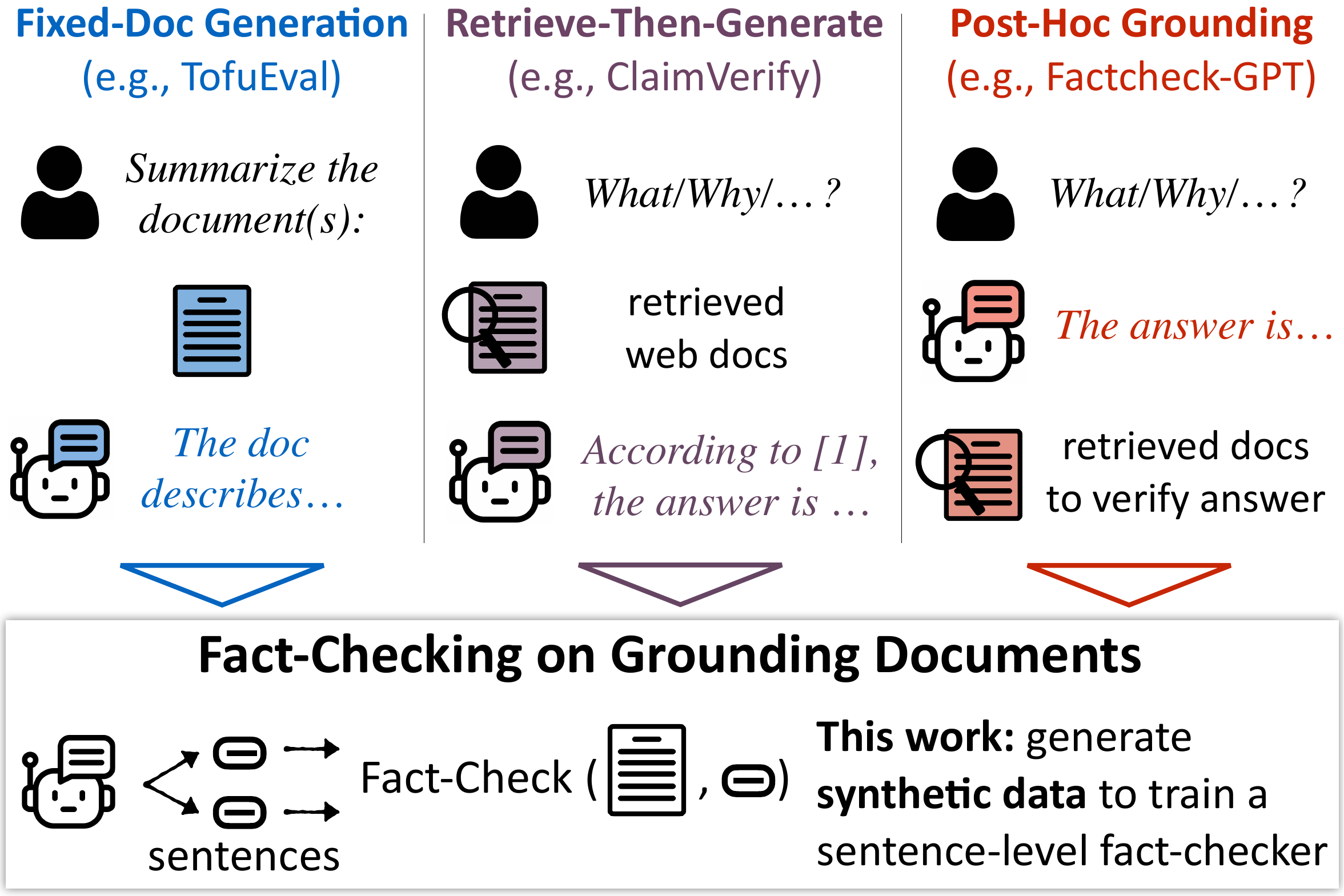}
\caption{We unify the task of fact-checking across various settings that rely on grounding documents. We train a small sentence-level fact-checker by leveraging new synthetically generated data, which demonstrates strong performance on a new unified benchmark LLM-\textsc{AggreFact}, comparable to GPT-4 but 400x cheaper.} \label{fig:main-fig}
\end{figure}

Freeform generation of responses is a flexible way to employ large language models (LLMs) for question answering, summarization, and beyond. However, this kind of generation can lead to factual errors, the ``hallucination'' problem in LLMs \cite{falke-etal-2019-ranking, maynez-etal-2020-faithfulness, mckenna-etal-2023-sources, zhang2023language}. Such errors arise in generation settings where an LLM is prompted closed-book, but its parametric knowledge may be insufficient to produce the right facts \cite{min-etal-2023-factscore, mallen-etal-2023-trust, zhou-etal-2023-context, chen2023understanding}. Different but related errors occur in grounded generation settings where evidence is already available, like summarization of input documents or retrieval-augmented question answering, where an LLM can blend information incorrectly \cite{liu-etal-2023-evaluating, tang2024tofueval}.

Past work has largely dealt with these problems separately. We can post-hoc verify closed-book generated answers by retrieving supporting documents and checking the answer against them \cite{gao-etal-2023-rarr, malaviya2024expertqa, jacovi2024chainofthought}, which requires precise checking as many statements are not exactly supported or may have conflicting information available \cite{Wang2023FactcheckGPTEF, Glockner2024}. When the grounding is known in settings like summarization, the attribution problem can be cleanly framed as document-level textual entailment \cite{nie-etal-2020-adversarial, yin-etal-2021-docnli}\, and has been studied extensively for smaller language models \cite{falke-etal-2019-ranking, goyal-durrett-2020-evaluating, goyal-durrett-2021-annotating, laban-etal-2022-summac, tang-etal-2023-understanding}.

The problems in this space have a shared primitive operation: the need to check a statement against grounding documents, either retrieval-augmented content or post-hoc retrieved evidence. We call this primitive \textbf{fact-checking on grounding documents}, shown in Figure~\ref{fig:main-fig}. Implementations of this primitive need to be accurate, spotting subtle errors while maintaining a low false positive rate, as most generated statements are correct. They also need to be efficient: a single LLM response may contain dozens of facts to verify, and self-verification with an LLM may increase cost by an order of magnitude \cite{weng-etal-2023-large, gero2023selfverification}. For instance, the 110-150 word biographies in FActScore \citep{min-etal-2023-factscore} contain 26-41 atomic facts that are checked against 5 documents each, resulting in 130-205 entailment checks. 

In this work, we build an efficient system for fact-checking on grounding documents. Our key insight is to develop a new synthetic training dataset which is tailored to the complexities of the fact-checking task. Unlike standard distillation from LLMs \cite{alpaca, hsieh-etal-2023-distilling}, this setting differs in that we do not necessary have access to task instances that we can label with strong LLMs. For instance, in datasets like ExpertQA \cite{malaviya2024expertqa}, even the inputs to the LLM are expert-written questions. As a result, we synthesize challenging fact-checking instances from the ground up, as a scalable way to teach a small model how to simultaneously verify multiple facts in a sentence against multiple sentences in grounding documents. Our system, MiniCheck, is an instance of Flan-T5 \cite{chung2022scaling} fine-tuned on this data plus standard entailment data \cite{nie-etal-2020-adversarial}.

For our experiments, we introduce a new unified benchmark, \benchmarkname{}, which aggregates 10 existing datasets for both closed-book and grounded generation settings. In each constituent dataset, sentence-level factual errors are labeled by human annotators. We show that MiniCheck can perform as well as GPT-4 in aggregate and substantially outperform past fine-tuned systems like AlignScore \cite{zha-etal-2023-alignscore}. Moreover, we find that decomposition of sentences into atomic facts, which has been explored in past work \cite{kamoi-etal-2023-wice, gao-etal-2023-rarr, Wang2023FactcheckGPTEF}, is not necessary to achieve this high performance.

Our contributions are as follows: (1) Two synthetic data generation methods to address the challenges of fact-checking on grounding documents. (2) A new benchmark unifying factual evaluation on closed-book and grounded generation settings. (3) Evaluation shows that our MiniCheck system can beat previous specialized systems by 4\% to 10\% in absolute values, despite using less fine-tuning data, and is on par with GPT-4 with a much smaller model size, faster inference speed, and 400 times less cost. Furthermore, we can do this without a separate claim decomposition step.

\section{Background and Motivation}
\label{sec:background}

\paragraph{Problem Setup: Claim Verification} We assume a collection of statements to be checked consisting of sentences $\mathbf{c} = [c_1,\ldots, c_{|\mathbf{c}|}]$. Typically, this will be a sequence of sentences produced by an LLM. Each sentence $c_i$ has an associated set of grounding documents $\mathcal{D}_i =\{D_{i,1},\ldots,D_{i,|\mathcal{D}_i|}\}$. These different $\mathcal{D}_i$ per sentence accommodate post-hoc retrieval settings where each sentence has different retrieved evidence; however, some settings may use shared evidence across all sentences or even a single grounding document for tasks like single-document summarization (i.e., all $\mathcal{D}_i$ only contain the document being summarized).

We view these sentences as \emph{claims}. Our goal in this work is to build a system that can validate each claim against the documents. Following \citet{laban-etal-2022-summac}, we define a discriminator 
$$M(D_{i,j}, c_i) \in \{0, 1\},$$
that classifies each claim  $c_i$ into \texttt{unsupported}, $0$, or \texttt{supported}, $1$, according to a provided document $D_{i,j}$.\footnote{Following past work \cite{kamoi-etal-2023-wice, sanyal2024minds}, we disregard the usual ``contradiction'' class from textual    entailment, as contradictions are rare in our benchmark.} 

This process makes two assumptions. First, we assume that \textbf{each supported claim can be validated against a single document}; that is, claims are ``atomic enough''. Our methodology can be generalized to handle claims supported by multiple documents by simply appending multiple documents in the context of $M$, but we did not find it necessary in any of the datasets we studied.

Second, we assume that \textbf{we can perform our entailment checks on each sentence $c_i$ on its own, without context $c_{<i}$.} In general, sentences do need context to be understood (e.g., most sentences starting with pronouns), but this can be resolved through the use of a \emph{decontextualization} step \cite{choi-etal-2021-decontextualization}. Section~\ref{sec:rethinking} examines whether such a step improves performance of our system.

We judge a sentence $c_i$ by taking $\max_j M(D_{i,j},c_i)$: it is supported if and only if there exists some document that supports it.

\begin{figure}
\centering
\includegraphics[width=0.9\linewidth, trim=0mm 0mm 00mm 0mm,clip]{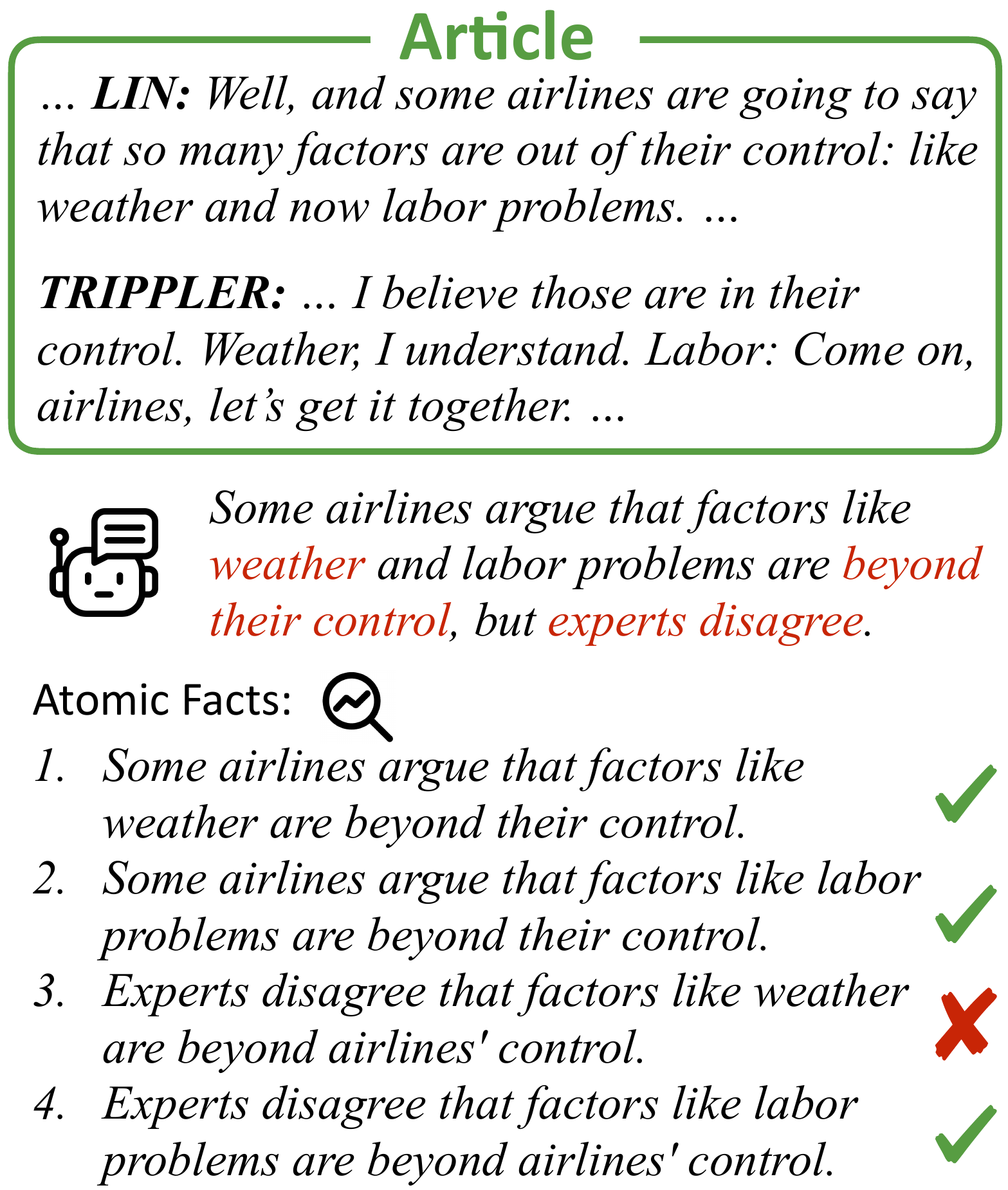}
\caption{An example dialogue snippet with an LLM-generated summary sentence, from the TofuEval dataset.} \label{fig:example-fig}
\end{figure}

\paragraph{Challenges of verification}

Two aspects of the task make this process challenging. First, there may be several individual facts in $D_{i,j}$ which are necessary to validate a claim $c_i$. For example, the LLM-generated sentence in Figure~\ref{fig:example-fig} can be broken down into four atomic facts. Each fact must be checked \emph{even if they are not explicitly materialized.}

Second, and relatedly, the claim, or a fact in the claim, may require making inferences that span multiple sentences within $D_{i,j}$. In Figure~\ref{fig:example-fig}, Lin initially argues that airlines have no control over either weather or labor issues. However, Trippler's later statement, \emph{``Weather, I understand. Labor: Come on, airlines, let's get it together,''}, implies agreement that weather is uncontrollable but suggests that labor problems are within the airlines' control. This indicates that the third atomic fact is unsupported by the document.

We argue that existing specialized fact-checkers fall short in effectively considering all atomic facts within a claim to be verified and struggle with reasoning across multiple sentences. Results in Appendix~\ref{sec:intrinsic-eval} support this characterization. To address this issue, we come up with two synthetic data generation methods (Section~\ref{sec:method}) to enhance the models' ability in these areas. We discuss the relation to prior work in Section~\ref{sec:related}. 

\section{Methodology: Training Data Synthesis} \label{sec:method}

To address these challenges, new data is required. Existing datasets like MNLI \cite{williams-etal-2018-broad} and ANLI \cite{nie-etal-2020-adversarial} do not feature instances that reflect the complexity of LLM fact-checking. Annotation of real errors is challenging to scale; datasets of such errors (including those in \benchmarkname{}) are largely test-only.

Our goal is to construct a dataset $\{(D_i, c_i, y_i)\}_{i=1}^N$ of $N$ instances of documents $D_i$ paired with claims $c_i$ with label $y_i \in \{0,1\}$, using two novel synthetic data generation methods (Figure~\ref{fig:data-gen-fig}). Statistics about our final synthetic training data can be found in Table~\ref{tab:train-data-stats}. A small-scale human evaluation of the synthetic data quality can be found in Appendix~\ref{sec:human-eval}. Additional details, including the sources of claims and documents and examples of generated data, can be found in Appendix~\ref{sec:synthetic-data-stats}. We provide all prompts and quality assurance details in Appendix~\ref{sec:prompts}.

\subsection{Claim to Doc (C2D) Generation} \label{sec:c2d}

In the C2D method, we assume that we have access to a set of human-written claim statements. The goal is to generate synthetic documents that require models be able to check multiple facts in the claim against multiple sentences each.

\paragraph{Step 1: Claim decomposition} Given a claim $c$, we first decompose it into a set of atomic facts $\mathbf{a}$ with GPT-3.5:
$\mathrm{\texttt{Decomp}}(c) = \{a_1, \ldots, a_l\}.$

\paragraph{Step 2: Atomic fact expansion} For the claim $c$, we ask GPT-4 to generate a pair of sentences for each of its atomic facts with a 4-shot prompt:
$$\mathrm{\texttt{SentPair}}(a_i) = (s_{i,1}, s_{i,2}), \forall i \in \{1, \ldots, l\}.$$
The generated sentence pairs are designed such that the atomic fact is supported if and only if the information from both sentences is combined.

\paragraph{Step 3: Supporting document generation} After expanding atomic facts $\mathbf{a}$ into sentences $\mathbf{s} = \{s_{1,1}, s_{1,2}, \ldots, s_{l,1}, s_{l,2}\}$, we ask GPT-4 to generate a document $D$ that mentions all sentences from the generated sentence pairs in its own words $D = \mathrm{\texttt{PassageGen}}(\mathbf{s})$
with a zero-shot prompt.\footnote{We ask GPT-4 to not state deduced facts or conclusions based on the provided sentences, and we find that GPT-4 can follow this instruction well.}

By following these steps, we create a triplet $(D, c, y=1)$. This procedure increases the difficulty of the task by ensuring that multiple-sentence reasoning is required to correctly classify a claim.

\begin{figure}
\centering
\includegraphics[width=\linewidth, trim=0mm 0mm 00mm 0mm,clip]{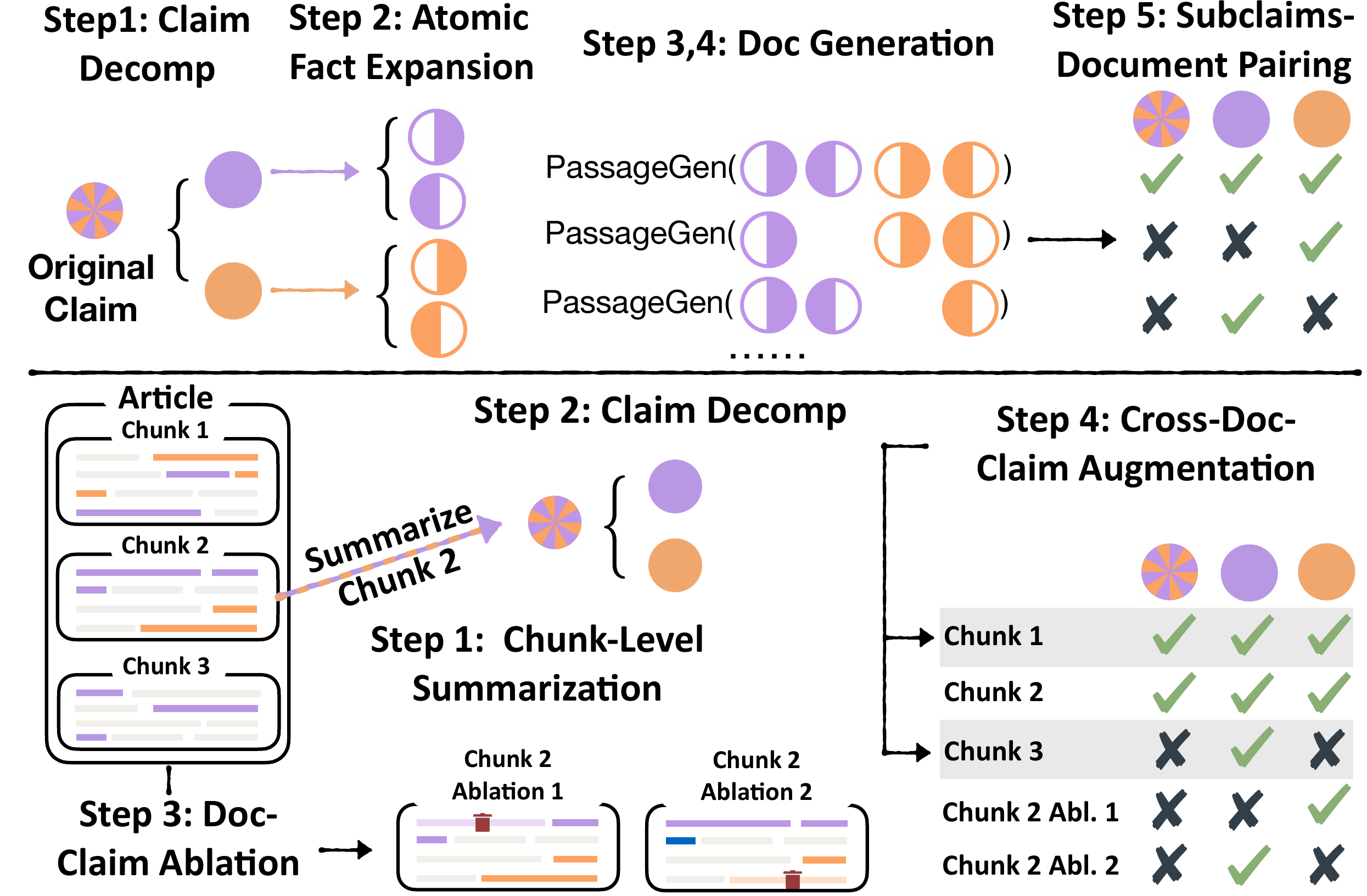}
\caption{Our synthetic data generation pipeline: C2D (upper) and D2C (lower). We illustrate with a claim that contains two atomic facts. Examples of generated data can be found in Appendix~\ref{sec:synthetic-data-stats}.} \label{fig:data-gen-fig}
\end{figure}

\paragraph{Step 4: Nonsupporting document generation} By construction, an atomic fact $a_i$ in the claim $c$ is supported by the sentence pair $(s_{i,1}, s_{i,2})$ mentioned in the generated document $D$. Therefore, by omitting one of the sentences from the pair in a newly generated document $D'$, it is likely that $a_i$, and consequently $c$, is no longer supported by $D'$ (except in cases of redundancy in the sentences $\mathbf{s}$).  More formally, we can construct a document $D'_{a_{i\setminus j}}$ that \emph{probably} cannot support fact $a_i$ in $c$ (and hence $c$) by removing sentence $s_{i, j}$ from its sentence pair:
$$D'_{a_{i\setminus j}} = \mathrm{\texttt{PassageGen}}(\mathbf{s} \setminus {s_{i, j}}),$$
for all $i \in \{1, \ldots, l\}$ and $j \in \{1, 2\}$ (Figure~\ref{fig:data-gen-fig}; top right). To collect documents that do not support the claim $c$, we retain $D'_{a_{i\setminus j}}$ if $a_i$ cannot be supported by the information combined from the remaining sentence from its sentence pair and other atomic facts ($s_{i, 3-j} \cup \{\mathbf{a} \setminus {a_i}\}$) via an entailment check by GPT-4. Note that this entailment check is again more accurate than directly checking $a_i$ against $D'_{a_{i\setminus j}}$ due to the shorter context.

\paragraph{Step 5: Pairing subclaims and generated documents} We have collected tuples $(D, c, 1)$ and $(D'_{a_{i\setminus j}}, c, 0)$ for some $i$ and $j$. We can further augment this data to produce more examples. We first generate a power set $\mathrm{\texttt{Power}}(\mathbf{a})$, that consists of all possible subsets of atomic facts $\mathbf{a}$ in $c$, but excludes the empty set. We then create a set of augmented subclaims $\mathrm{Aug}(c)$ by merging atomic facts from each subset:
$$\mathrm{Aug}(c) = \{\mathrm{\texttt{Merge}}(\mathbf{a'}): \forall \mathbf{a'} \in \mathrm{\texttt{Power}}(\mathbf{a})\}.$$
It follows that we obtain tuples $(D, c', 1)$ for every $c' \in \mathrm{Aug}(c)$. Similarly, for each $D'_{a_{i\setminus j}}$, we generate tuples $(D'_{a_{i\setminus j}}, \mathrm{\texttt{Merge}}(\mathbf{a'}), 1)$ if $a_i \notin {\mathbf{a'}}$, indicating that the document still supports the subclaim absent the atomic fact $a_i$. Conversely, we have $(D'_{a_{i\setminus j}}, \mathrm{\texttt{Merge}}(\mathbf{a'}), 0)$ if $a_i \in {\mathbf{a'}}$, suggesting that the document does not support the subclaim due to the absence of $a_i$.

Because the same subclaim is supported by certain documents and unsupported by others depending on the presence or absence of specific atomic facts, we achieve the same benefits that training on contrast sets provides \cite{cao-wang-2021-cliff, liu-etal-2022-brio, tang-etal-2023-less}, namely making the model more sensitive to the specifics of the decision boundary and encouraging it to consider all atomic facts within a claim during prediction.

\subsection{Doc to Claim (D2C) Generation} \label{sec:d2c}

In the D2C method, our objective is to enhance the diversity of documents and ensure that the documents are more realistic than those in C2D, thereby reducing the distribution shift between synthetic documents used during training and real documents at test time. To achieve this, we assume that we have access to a set of human-written documents to start with. The goal is to generate claims and pair them with portions of these human written documents, which, once again, require multi-sentence, multi-fact reasoning to check the claims.

\paragraph{Step 1: Chunk-level summarization} We first divide a human written document into three chunks $\{D_1, D_2, D_3\}$ with approximately equal length. We then use GPT-4 to generate a summary sentence for each chunk, resulting in a set of summary sentences $\mathbf{c} = \{c_1, c_2, c_3\}$. We assume these generated summary sentences are factually consistent with respect to their corresponding chunks, \emph{i.e.} $(D_i, c_i, 1)$ for all $i$, as each chunk is short and LLMs can almost always generate factual summaries in this setting \cite{Zhang2024}.

\paragraph{Step 2: Claim decomposition and subclaim augmentation} Similar to the C2D method, for a summary sentence $c_i$ in $\mathbf{c}$, we decompose it into atomic facts $\mathbf{a_i} = \{a_{i,1}, \ldots, a_{i, l}\}$, and create a set of augmented subclaims $\mathrm{Aug}(c_i) = \{\mathrm{\texttt{Merge}}(\mathbf{a}'_{i}): \forall \mathbf{a}'_{i} \in \mathrm{\texttt{Power}}(\mathbf{a}_i)\}$.

\paragraph{Step 3: Document-claim augmentation} This step aims to do data augmentation on a $(D_i, c_i)$ pair. Given a chunk $D_i = \mathrm{\texttt{Concat}(\mathbf{s})}$, which is the concatenation of $n$ sentences $\mathbf{s} = \{s_{i,1}, ..., s_{i,n}\}$, we construct new documents by iteratively removing each sentence $s_{i,j}$ from $\mathbf{s}$:
$$D'_{i \setminus j} = \mathrm{\texttt{Concat}}(\mathbf{s}\setminus \{s_{i,j}\}).$$
We then determine the entailment label for each atomic fact $a_{i, k}$ in $c_i$, where $k \in \{1, \ldots, l\}$:
$$L^{-j}(a_{i, k}) = \mathrm{Ent}(D'_{i \setminus j}, a_{i, k}) \in \{0, 1\}.$$
Similar to step 5 in C2D, if $L^{-j}(a_{i, k}) = 1$ for all $a_{i, k} \in \mathbf{a}'_i$, we create tuples $(D'_{i \setminus j}, \mathrm{\texttt{Merge}}(\mathbf{a}'_i), 1)$. Conversely, if there exists any $a_{i, k} \in \mathbf{a}'_i$ such that $L^{-j}(a_{i, k}) = 0$, we then create tuples $(D'_{i \setminus j}, \mathrm{\texttt{Merge}}(\mathbf{a}'_i), 0)$.

\paragraph{Step 4: Cross-document-claim augmentation} The objective of this step is to perform data augmentation on a $(D_j, c_i)$ pair, where $j \neq i$. The rationale behind this is that the important information in a document can be conveyed multiple times in various ways. Given that each chunk $D_i$ has an associated summary $c_i$, it is probable that the summary $c_i$ conveys some information that can be indirectly supported by other chunks $D_j$ within the document, even if $D_j$ are not used to generate $c_i$. Therefore, we consider chunks $D_j$, where $j \neq i$, as more challenging chunks to either support or refute the claim $c_i$ or its atomic facts $\mathbf{a}_i$.

More formally, we determine the entailment label for each atomic fact $a_{i, k}$ in $c_i$, using the document chunk $D_j$, where $k \in \{1, \ldots, l\}$, and $j \neq i$:
$$L^{D_j}(a_{i, k}) = \mathrm{Ent}(D_j, a_{i, k}) \in \{0, 1\}.$$    
If $L^{D_j}(a_{i, k}) = 1$ for all $a_{i, k} \in \mathbf{a}'_i$, we create tuples $(D_j, \mathrm{\texttt{Merge}}(\mathbf{a}'_i), 1)$. Conversely, if there exists any $a_{i, k} \in \mathbf{a}'_i$ such that $L^{D_j}(a_{i, k}) = 0$, we then create tuples $(D_j, \mathrm{\texttt{Merge}}(\mathbf{a}'_i), 0)$.

\begin{table}
\small
\centering
\renewcommand{\tabcolsep}{1.75mm}
\begin{tabular}{ccccccc}
\toprule
\textbf{Data} &
  \textbf{Size} &
  \begin{tabular}[c]{@{}c@{}}\textbf{Uniq.}\\ \textbf{Claim}\end{tabular} &
  \begin{tabular}[c]{@{}c@{}}\textbf{Uniq.}\\ \textbf{Doc}\end{tabular} &
  \begin{tabular}[c]{@{}c@{}}\textbf{Doc}\\ \textbf{Len}\end{tabular} &
  \begin{tabular}[c]{@{}c@{}}\textbf{Claim}\\ \textbf{Len}\end{tabular} &
  \begin{tabular}[c]{@{}c@{}}\textbf{\% of}\\ \textbf{Neg}\end{tabular} \\
\midrule
C2D &
  7076 &
  2004 &
  1188 &
  189 &
  19 &
  42\% \\
D2C &
  7319 &
  1392 &
  4967 &
  164 &
  12 &
  65\% \\
\bottomrule
\end{tabular}
\caption{\textbf{Statistics of synthetic training data.} 
Amount of synthetic data for training, the number of unique claims and documents, the average number of words in documents and claims, and the proportion of unsupported claims.} \label{tab:train-data-stats}
\end{table}

\subsection{\textsc{MiniCheck} Models} \label{sec:our-models}

We fine-tune three models with various model architectures by leveraging our synthetic data. We use the standard cross-entropy loss for all models. See Appendix~\ref{sec:implement-detail} for training details.

\paragraph{MiniCheck-\textsc{Dbta} and MiniCheck-\textsc{FT5}} As models trained on the ANLI dataset \cite{nie-etal-2020-adversarial} have demonstrated strong performance \cite{kamoi-etal-2023-wice, honovich-etal-2022-true}, we integrate our data with the ANLI dataset for fine-tuning \texttt{deberta-v3-large} \cite{he2021debertav3} and \texttt{flan-t5-large} \cite{chung2022scaling}. We take a subset (21K) of the ANLI training data, selecting examples where their trained entailment models made incorrect predictions during dataset construction. Training on more of ANLI was not effective.

Combining these 21K datapoints with our 14K-sized dataset, we have 35K training datapoints in total. We map the labels \texttt{contradiction} and \texttt{neutral} from ANLI to \texttt{unsupported}. 

\paragraph{MiniCheck-\textsc{Rbta}} We also explore whether it is possible to improve upon the previous AlignScore \cite{zha-etal-2023-alignscore} system, the existing SOTA specialized fact-checking model. We fine-tune the tuned \texttt{roberta-large} \cite{liu2019roberta} model from AlignScore with a binary classification head, on our 14K synthetic datapoints.

\paragraph{Producing classification decisions} Although our task is framed as binary classification, in reality the models we have are of the form $M(D_i, c_i) \rightarrow z \in \mathbb{R}$, mapping each (document, claim) pair to a score in the range $z \in [v_{\mathrm{min}}, v_{\mathrm{max}}]$. Following \citet{laban-etal-2022-summac, zha-etal-2023-alignscore, tang-etal-2023-understanding}, we convert each method into a binary classifier $M(D_i, c_i) \to \{0, 1\}$ by picking a threshold $t$ such that we predict 1 if $M(D_i, c_i) > t$ and 0 otherwise. Unless otherwise specified, we set $t=0.5$.

\section{\benchmarkname{} Benchmark} \label{sec:benchmark}

We construct a fact verification benchmark, \benchmarkname{}, by aggregating 10 of the most up-to-date publicly available datasets on factual consistency evaluation across both closed-book and grounded generation settings.

\paragraph{Characteristics} In \benchmarkname{}, all datasets contain human-annotated \texttt{(document, claim, label)} tuples. The documents come from diverse sources, including Wikipedia paragraphs, interviews, web text, covering domains such as news, dialogue, science, and healthcare. The claims to be verified are mostly generated from recent generative models (except for one dataset of human-written claims), without any human intervention in any format, such as injecting certain error types into model-generated claims. An overview of the \benchmarkname{} is shown in Figure~\ref{tab:benchmark-overview}, with statistics and detailed dataset descriptions in Appendix ~\ref{sec:benchmark-stats}.

\begin{figure}[t!]
\centering
\includegraphics[width=1\linewidth, trim=0mm 0mm 00mm 0mm,clip]{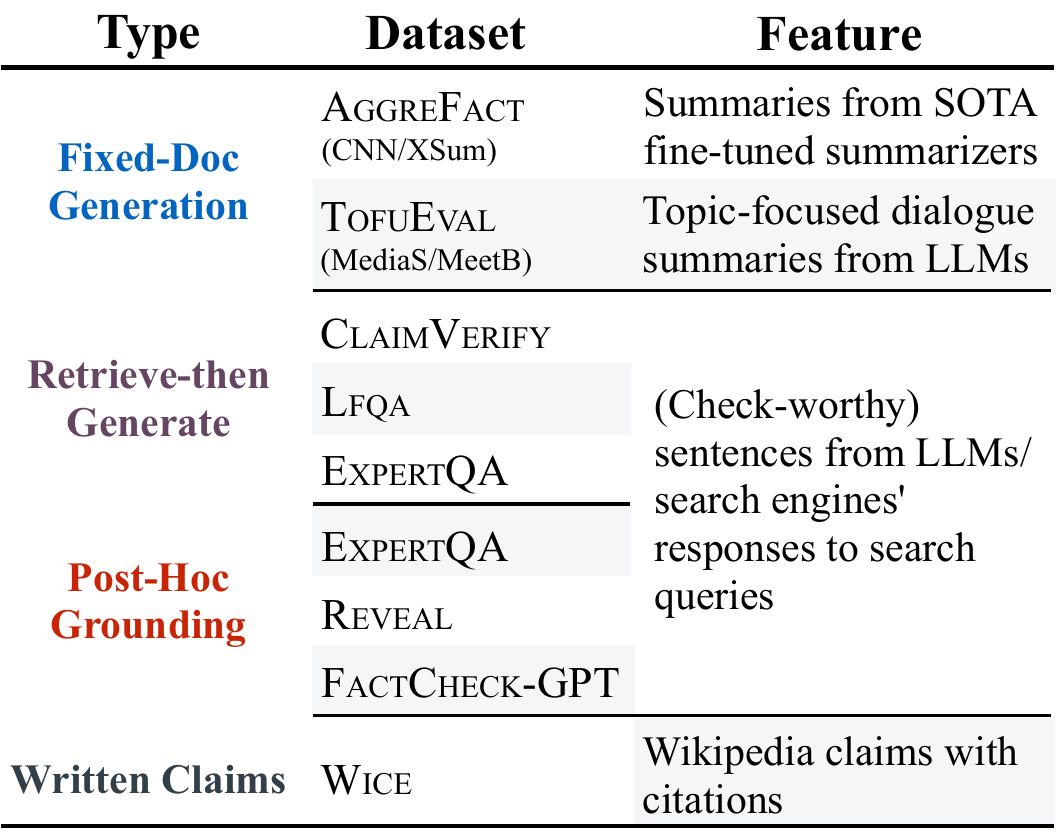}
\caption{10 datasets in \benchmarkname{}. Details of these datasets as well as related but excluded datasets can be found in Appendix~\ref{sec:benchmark-stats}.} \label{tab:benchmark-overview}
\end{figure}

\subsection{Benchmark Details}

\paragraph{Validation/Test set split} For the datasets from \textsc{AggreFact} and \textsc{TofuEval}, as well as \textsc{Wice} and \textsc{ClaimVerify}, we directly use the existing validation and test splits from the original work. For \textsc{Reveal}, \textsc{FactCheck-GPT},  \textsc{ExpertQA} and \textsc{Lfqa}, we randomly divide each of them into validation and test sets (50\%/50\%), assuring that responses to unique queries do not appear in both sets.

One potential use of the validation data is to allow for \textbf{per-dataset threshold tuning}. This setting is used in substantial past work \citep{laban-etal-2022-summac, luo2023chatgpt, zha-etal-2023-alignscore, tang-etal-2023-understanding, tang2024tofueval}. However, we do not follow this trend in order to focus on building systems that can be deployed zero-shot across multiple downstream tasks, without any additional hyperparameter tuning. Instead, for $M(d, c) \to z \in [v_{\mathrm{min}}, v_{\mathrm{max}}]$, the threshold is set as the midpoint of the output score range $t = (v_{\mathrm{min}} + v_{\mathrm{max}}) / 2$, which is 0.5 for most fact-checkers. In practice, most fact-checkers return scores at the extremes of the range, so small tweaks on this procedure have little effect. See Appendix~\ref{sec:result-tuning-thresh} for the results of the threshold tuning setting, which yields qualitatively similar results.

\begin{table*}[t]
\small
\centering
\renewcommand{\tabcolsep}{1.5mm}
\begin{tabular}{lcccccccccc|c}
\toprule
           & \multicolumn{10}{c}{\textbf{\benchmarkname{}} (\emph{without} threshold tuning)}                                                     \\
\cmidrule(r){2-11} 
\multirow{2}{*}{\begin{tabular}[c]{@{}l@{}}\textbf{Model}\\ \textbf{Name}\end{tabular}} &
  \multicolumn{2}{c}{\textbf{\textsc{AggreFact}}} &
  \multicolumn{2}{c}{\textbf{\textsc{TofuEval}}} &
  \multirow{2}{*}{\textbf{\textsc{Wice}}} &
  \multirow{2}{*}{\textbf{\textsc{Reveal}}} &
  \multirow{2}{*}{\begin{tabular}[c]{@{}c@{}}\textbf{\textsc{Claim}}\\ \textbf{\textsc{Verify}}\end{tabular}} &
  \multirow{2}{*}{\begin{tabular}[c]{@{}c@{}}\textbf{\textsc{Fact}}\\ \textbf{\textsc{Check}}\end{tabular}} &
  \multirow{2}{*}{\begin{tabular}[c]{@{}c@{}}\textbf{\textsc{Expert}}\\ \textbf{\textsc{QA}}\end{tabular}} &
  \multirow{2}{*}{\textbf{\textsc{Lfqa}}} &
  \multirow{2}{*}{\textbf{Avg}} \\
\cmidrule(r){2-3} \cmidrule(r){4-5} 

           & \textbf{CNN}  & \textbf{XSum} & \textbf{MediaS} & \textbf{MeetB} &      &      &      &      &      &    &  \\
\midrule
Gemini-Pro & 49.4 & 60.6 & 63.8 &  65.8  & 65.8 & 85.5 & 61.8 & 76.8 & 56.8 & 75.9 & 66.2 \\
PaLM2-Bison & 52.4 & 59.0 & 68.3 & 73.6 & 63.4 & 84.2 & 60.5 & 76.4 & 56.6 & 71.4 & 66.6 \\
Mistral-8x7B & 55.0 & 65.5 & 68.5 & 73.3 & 63.8 & 80.8 & 64.3 & 75.1 & 56.3 & 70.8 & 67.3  \\
GPT-3.5 & 63.2 & 72.4 & 66.8 & 73.4 & 68.5 & 84.7 & 65.2 & 70.8 & 57.2 & 73.8 & 69.6 \\
Claude-2.1 & 59.9 & 66.4 & 69.2 & 72.3 & 64.3 & \cellcolor{green2}88.2 & 69.7 & 79.3 & 59.8 & 78.2 & 70.7 \\
Mistral-Large & 58.4 & \cellcolor{green2}76.3 &  67.3 & 78.9  & 76.6 & \cellcolor{green2}88.4 & 67.6 & 79.0 & \cellcolor{green2}60.0 & 81.7 & 73.4 \\
Claude-3 Opus & \cellcolor{green2}65.2 & \cellcolor{green2}72.4 & \cellcolor{green2}74.1 & \cellcolor{green2}82.4 & 75.0 & 83.8 & 69.3 & \cellcolor{green2}78.8 & \cellcolor{green2}58.8 & 81.6 & 74.1 \\
GPT-4   & \cellcolor{green2}66.7 & \cellcolor{green2}76.5 & \cellcolor{green2}71.4 & 79.9 & \cellcolor{green2}80.4 & \cellcolor{green2}87.8 & 67.6 & \cellcolor{green2}79.9 & \cellcolor{green2}59.2 & 83.1 &  \cellcolor{green2}75.3 \\
\midrule
SummaC-CV   & \cellcolor{green2}65.2 & 54.5 & 63.7 & 62.8 & 54.3 & 67.7 & 70.9 & 53.4 & 54.9 & 62.1 & 62.1 \\
T5-NLI-Mixed & 54.6 &  52.3 & 59.1 & 55.3 & 55.3 & \cellcolor{green2}87.2 & 59.5 & 69.0 & 55.6 & 61.8 & 61.0 \\
FT5-ANLI-L & 51.2 &  60.0 & 57.4 & 60.1 & 67.0 & 77.5 & 58.3 & 67.7 & 52.2 & 63.0 & 61.4 \\
DAE   & 50.8 & 59.1 & 65.1 & 69.5 & 58.5 & 81.3 & 64.0 & 72.5 & 56.2 & 72.2 & 64.9 \\
QAFactEval & 54.3 & 62.1 & 61.3 & 65.7 & 62.5 & 83.2 & 73.2 & 66.1 & 56.0 & 80.6 & 66.5 \\
SummaC-ZS   & 51.1 & 61.5 & 69.5 & 71.0 & 62.8 & 85.3 & 69.7 & 75.2 & 55.2 & 77.6 & 67.9 \\
AlignScore & 52.4 & 71.4 & 69.2 & 72.6 & 66.0 & 85.3 & 69.6 & 74.3 & \cellcolor{green2}58.3 & \cellcolor{green2}84.5 & 70.4 \\
\midrule
MiniCheck-\textsc{Dbta}  & \cellcolor{green2}64.2 & 71.0 & 69.3 & 72.7 & 69.4 & \cellcolor{green2}87.3 & 75.6 & 73.0 & \cellcolor{green2}58.9 & \cellcolor{green2}83.9 & 72.6 \\
MiniCheck-\textsc{Rbta}  & \cellcolor{green2}63.7  & 70.8 & \cellcolor{green2}71.9 & 75.9 & 67.6 & \cellcolor{green2}88.8 & \cellcolor{green2}77.4 & 73.3 & 57.4 & \cellcolor{green2}84.4 & 72.7 \\
MiniCheck-\textsc{FT5}    & \cellcolor{green2}69.9 & \cellcolor{green2}74.3 & \cellcolor{green2}73.6 & 77.3 & 72.2 & 86.2 & \cellcolor{green2}74.6 & 74.7 & \cellcolor{green2}59.0 & \cellcolor{green2}85.2 & \cellcolor{green2}74.7 \\
\bottomrule
\end{tabular}
\caption{Performance (BAcc) of models on the test set of \benchmarkname{} without per-dataset threshold tuning. Models are split into \emph{LLM-based fact-checkers | specialized fact-checkers | Ours}. We highlight the \colorbox{green2}{best} performance for each dataset, where multiple green highlights indicate systems indistinguishable from the best according to a paired bootstrap test with 1000 runs and p-value < 0.05. Details for -Dectx and -Decmp are in Section~\ref{sec:rethinking}. Our MiniCheck models outperform other specialized evaluators and MiniCheck-\textsc{FT5} reaches the performance of GPT-4.} \label{tab:result-zs}
\end{table*}

\paragraph{Evaluation Metric} Following \citet{laban-etal-2022-summac, fabbri-etal-2022-qafacteval, tang-etal-2023-understanding}, we evaluate the performance of fact-checkers using balanced accuracy (BAcc):
$\mathrm{BAcc} = \frac{1}{2}\left(\frac{\mathrm{TP}}{\mathrm{TP + FN}} + \frac{\mathrm{TN}}{\mathrm{TN + FP}}\right)$,
where TP, TN, FP, and FN represent true/false positives/negatives.

\section{Experimental Setup} \label{sec:exp-setup}

We include the following specialized fact-checkers: \textbf{T5-NLI-Mixed} \cite{honovich-etal-2022-true}, \textbf{DAE} \cite{goyal-durrett-2021-annotating}, \textbf{QAFactEval} \cite{fabbri-etal-2022-qafacteval}, \textbf{SummaC-ZS} and \textbf{SummaC-CV} \cite{laban-etal-2022-summac}, \textbf{AlignScore} \cite{zha-etal-2023-alignscore}, and \textbf{FT5-ANLI-L} that fine-tunes \texttt{flan-t5-large} on the full ANLI training set. A meta-comparison of those specialized fact-checkers and our models can be found in Table~\ref{tab:model-card}. More inference details can be found in Appendix~\ref{sec:evaluators-detail}. 

We also include the following LLMs as fact-checkers: \textbf{Gemini-Pro} \cite{geminiteam2023gemini}, \textbf{PaLM2-Bison} \cite{Thoppilan2022LaMDALM}, \textbf{Mistral-8x7B}, \textbf{Mistral-Large} \cite{jiang2024mixtral}, \textbf{Claude 2.1}, \textbf{Claude 3 Opus} \cite{bai2022constitutional}, \textbf{GPT-3.5} and \textbf{GPT-4} \cite{OpenAI2023GPT4TR}. More details about the models can be found in Appendix~\ref{sec:llm-info}. For the LLM-based fact-checkers, we adapt a prompt from \citet{luo2023chatgpt} for zero-shot prediction, which can be found in Appendix~\ref{sec:prompts}.

\section{Results} \label{sec:results}

\subsection{Main Results}

\paragraph{Our synthetic data improves performance across diverse model architectures.} Table~\ref{tab:result-zs} demonstrates that our synthetic data gives strong performance when used in three different backbone models: RoBERTa, DeBERTa, and Flan-T5. These models outperform prior models of a similar scale. Notably, MiniCheck-\textsc{FT5} achieves a 4.3\% overall improvement over AlignScore, outperforming it on 6 out of 10 datasets and matching its performance on the remaining 4. We attribute its additional 2\% gain over MiniCheck-\textsc{Rbta} and -\textsc{Dbta} to its larger model size. However, model size alone does not guarantee superior performance, as evidenced by T5-NLI-Mixed and FT5-ANLI-L, which, despite being trained on NLI datasets, underperform on most of the benchmark settings. This underscores the importance of training data selection in addition to model capacity.

\paragraph{Our models achieve performance on par with the most capable LLM-based fact-checkers.} In the top rows of Table~\ref{tab:result-zs}, we present the performance of strong LLM-based fact-checkers. We observe that existing specialized fact-checkers achieve similar performance to non-frontier LLM-based fact-checkers like Mistral-8x7B and GPT-3.5. MiniCheck-\textsc{Rbta} and MiniCheck-\textsc{Dbta} can surpass these non-frontier LLM-based fact-checkers by a large margin. MiniCheck-\textsc{FT5} achieves the same performance as Claude-3 Opus and is close to GPT-4, but with a much smaller model size.

\begin{table}[t!]
\small
\centering
\renewcommand{\tabcolsep}{1.3mm}
\begin{tabular}{lcccc}
\toprule
\multirow{2}{*}{\begin{tabular}[c]{@{}l@{}}\textbf{Model}\\ \textbf{Name}\end{tabular}} &
  \multirow{2}{*}{\begin{tabular}[c]{@{}c@{}}\textbf{Backbone}\\ \textbf{Model}\end{tabular}} &
  \multirow{2}{*}{\begin{tabular}[c]{@{}c@{}}\textbf{Model}\\ \textbf{Size}\end{tabular}} &
  \multirow{2}{*}{\begin{tabular}[c]{@{}c@{}}\textbf{\# FT}\\ \textbf{Data}\end{tabular}} &
  \multirow{2}{*}{\begin{tabular}[c]{@{}c@{}}\textbf{Cost}\\ \textbf{(\$)}\end{tabular}} \\
            &                               &                      &      &                        \\
\midrule
T5-NLI-Mixed & T5-XXL      & 11B        &  1,697K       &  7.39 \\ 
FT5-ANLI-L &  Flan-T5-L    & 770M        &  163K       &  0.24 \\
DAE         &   ELECTRA-B       &  110M         &  95K       &   0.26  \\
QAFactEval  & multiple$^*$    &  1.4B    &   -     &   1.87            \\
SummaC-ZS    & ALBERT-XL    &   60M   & 371K        &   0.85       \\
SummaC-CV    & ALBERT-XL   &    60M    & 381K       &   0.85     \\
AlignScore  & RoBERTa-L   &    355M    & 4,700K    &   0.20    \\
\midrule
MiniCheck-\textsc{Rbta}       & AlignScore   &    355M    & 14K    &  0.20     \\
MiniCheck-\textsc{Dbta}      & DeBERTa-L   &    355M    & 35K       &   0.20      \\
MiniCheck-\textsc{FT5}      & Flan-T5-L   &    770M    & 35K       &   0.24    \\
\bottomrule
\end{tabular}
\caption{Comparison of specialized fact-checkers on model sizes, training data sizes, and the inference cost (\$0.8/GPU-hr) on the 13K \benchmarkname{} test set. $^*$QAFactEval contains several model components, which sum up to 1.4B in size.} \label{tab:model-card}
\end{table}

\paragraph{Extended Analysis} See Appendix~\ref{sec:analysis} for an intrinsic evaluation on our synthethic data and an ablation study on our best model MiniCheck-\textsc{FT5}.

\subsection{Computational Cost Comparison}

We compare the computational cost of specialized fact-checkers and LLMs on the test set of \benchmarkname{}. For specialized fact-checkers, we use our own hardware and convert the prediction time on our GPUs to the equivalent cost of using cloud computing services (see Appendix~\ref{sec:machine-config} for details). For LLM-based fact-checkers, we compute the costs of corresponding API calls. Results are shown in Table~\ref{tab:model-card} and~\ref{tab:llm-throughput}. We see that specialized models in general have much lower inference costs. In particular, our most capable model MiniCheck-\textsc{FT5} has almost the same performance as GPT-4 but is more than 400 times cheaper.

\section{Rethinking LLM Fact-Checking}
\label{sec:rethinking}

We now revisit two other stages of the typical LLM fact-checking pipeline: claim decomposition and decontextualization. Surprisingly, we find that claim decomposition is not needed in our settings, contradicting prior work \cite{yang-zhu-2021-exploring-decomposition, kamoi-etal-2023-wice}. Furthermore, we find that decontextualization doesn't help on our benchmark, although we believe that it is needed in general.

\subsection{Claim Decomposition} \label{sec:claim-decmp}

We also experiment with a setting using claim decomposition. In this setting, we decompose each claim $c_i$ into atomic facts $\mathbf{a}_i$ with the prompt from \citet{kamoi-etal-2023-wice} and use a fact-checker to predict the factuality label for each $(D_i, a_{i, k})$ pair, $k \in \{1, \ldots, l\}$. If all atomic facts are supported by the document, then the claim is supported, and unsupported otherwise. We do this for every dataset except FactCheck-GPT which is already atomic facts. There are typically 2-4 atomic facts per claim across datasets.

\begin{table}
\small
\centering
\renewcommand{\tabcolsep}{1.2mm}
\begin{tabular}{lc|lc}
\toprule
\begin{tabular}[c]{@{}l@{}}\textbf{Model}\\ \textbf{Name}\end{tabular} &
  \begin{tabular}[c]{@{}c@{}}\textbf{Cost}\\ \textbf{(\$)}\end{tabular} &
  \begin{tabular}[c]{@{}l@{}}\textbf{Model}\\ \textbf{Name}\end{tabular} &
  \begin{tabular}[c]{@{}c@{}}\textbf{Cost}\\ \textbf{(\$)}\end{tabular} \\
\midrule
Gemini-Pro         & 5.24                   &  Claude-2.1         & 89.9\\
PaLM2-Bison             & 10.9       & Claude-3 Opus   & 165                \\
Mistral-8x7B           & 7.78                   &   GPT-3.5          & 4.75                  \\
Mistral-Large           & 90.2                        & GPT-4              & 107                 \\
GPT-4-Dectx & 161 & GPT-4-Decmp & 212 \\
\bottomrule
\end{tabular}
\caption{Inference cost comparison for API models on the 13K \benchmarkname{} test set. Both decontextualization and decomposition add cost to GPT-4. Overall, decoding our test set with the most capable models incurs significant cost.} \label{tab:llm-throughput}
\end{table}

We show the results from GPT-4 and a subset of specialized fact-checkers in Table~\ref{tab:decompose}. We observe near-zero performance change for GPT-4 and mixed changes for specialized fact-checkers. \textbf{Overall, there is no clear indication that decomposing claims into atomic facts can consistently improve models' performance.} Because this approach increases the inference time and costs by a factor of 2-4 for different datasets, depending on the average number of atomic facts per claim, we believe it should not be used until it provides a clear accuracy benefit.\footnote{Note that for Factcheck-GPT, retrieval operates over individual atomic facts. Decomposition may still be necessary to \emph{retrieve} the relevant information, but our results show that it may not be necessary for \emph{entailment checks}.} 

\begin{table}
\small
\centering
\renewcommand{\tabcolsep}{1.25mm}
\begin{tabular}{lcc}
\toprule
\textbf{Model} & \textbf{Decomposition} & \textbf{Decontextualization} \\
\midrule
GPT-4      & 75.6  \color{greenarrow}$(\boldsymbol{\uparrow 0.3})$ & 75.3  \color{black}$(+ \boldsymbol{0.0})$\\
\midrule
SummaC-CV      &   58.8 \color{red3}$(\boldsymbol{\downarrow 3.3})$  & 60.8 \color{red3}$(\boldsymbol{\downarrow 1.3})$\\
QAFactEval      &  64.6 \color{red3}$(\boldsymbol{\downarrow 1.9})$   & 66.4 \color{red3}$(\boldsymbol{\downarrow 0.1})$\\
SummaC-ZS     &   69.1 \color{greenarrow}$(\boldsymbol{\uparrow 1.2})$  & 67.7 \color{red3}$(\boldsymbol{\downarrow 0.2})$\\
AlignScore     &   71.5 \color{greenarrow}$(\boldsymbol{\uparrow 1.1})$  & 70.4 \color{black}$(+ \boldsymbol{0.0})$\\
\midrule
MiniCheck-\textsc{Rbta} &   73.2 \color{greenarrow}$(\boldsymbol{\uparrow 0.5}$)  &  72.4 \color{red3}$(\boldsymbol{\downarrow 0.3})$ \\
MiniCheck-\textsc{Dbta} &   72.7 \color{greenarrow}$(\boldsymbol{\uparrow 0.1})$  & 71.2 \color{red3}$(\boldsymbol{\downarrow 1.4})$ \\
MiniCheck-\textsc{FT5} &   73.3 \color{red3}$(\boldsymbol{\downarrow 1.4})$  & 74.1 \color{red3}$(\boldsymbol{\downarrow 0.6})$\\
\bottomrule
\end{tabular}
\caption{Average performance on the test set of \benchmarkname{} by aggregating predictions on decomposed claims (left); doing claim decontextualization where it is applicable (right). We show the performance change compared to predicting using original claims from Table~\ref{tab:result-zs}. Full results in Appendix Tables~\ref{tab:decompose-full} and \ref{tab:dectx-full}.} \label{tab:decompose}
\end{table}

\subsection{Claim Decontextualization}

As mentioned in Section~\ref{sec:background}, our approach relies on being able to check each sentence in isolation. However, phenomena like coreference and ellipsis may make sentences difficult to ground out of context. We can address this with an explicit \emph{decontextualization} step \cite{choi-etal-2021-decontextualization, Wang2023FactcheckGPTEF, jacovi2024chainofthought}. We experiment with \textsc{TofuEval}-MediaS, \textsc{TofuEval}-MeetB, \textsc{Wice}, \textsc{Reveal}, \textsc{ClaimVerify}, \textsc{ExpertQA} and \textsc{Lfqa}, which are the datasets in our benchmark where sentences need to be interpreted in context (\textsc{FactCheck} is already decontextualizaed). We prompt GPT-4 for decontextualization as shown in Appendix~\ref{sec:prompts}, using the previous claims or response sentences as context to expand the claim. Respectively, 33\%, 33\%, 39\%, 11\%, 35\%, 47\%, and 57\% of the claims from those datasets are changed after decontextualization.

In Table~\ref{tab:decompose}, we show the average fact-checking performance when using this decontextualization step (see the prompt in Table~\ref{tab:sent-decontext}). \textbf{These results suggest that models may make decent guesses about context-dependent content, particularly when the retrieval stage already implicitly enforces shared context between the claim and the grounding documents.} However, for tasks such as retrieval-augmented generation, we believe decontextualization still plays a crucial role in ensuring meaningful document retrieval. Furthermore, as LLMs scale further and their responses get more complex, the level of contextualization they feature may be higher, making this step more necessary.

\section{Related Work}
\label{sec:related}

\paragraph{Hallucinations in LLMs}

LLMs are prone to hallucinations across various settings \cite{huang2023survey, zhang2023siren, rawte2023survey}, generating information that cannot be supported by any source. For example, in the closed-book setting, where LLMs rely solely on their parametric knowledge, they may fabricate details when describing biographies or providing Wikipedia entity information \cite{min-etal-2023-factscore, guan2023language, mallen-etal-2023-trust}. In retrieval-augmented settings, where models have access to external documents to provide responses to user queries, they may generate supplementary information that is not faithful to the provided documents \cite{chiesurin-etal-2023-dangers, adlakha2023evaluating, Chen2024benchRAG}. Even when LLMs are provided with gold documents, such as in text summarization and simplification tasks, they still generate factually inconsistent outputs with diverse error types across different domains \cite{joseph-etal-2023-multilingual, shaib-etal-2023-summarizing, tang2024tofueval, Tang2023}. In this work, we construct a new benchmark dataset, \benchmarkname{}, which unifies human-annotated model responses across all settings, and evaluate the performance of existing fact-checkers and our proposed ones on the benchmark in detecting such errors.

\paragraph{Methods in Detecting Hallucinations}

When documents are directly available for model-generated sentences, such as in text summarization \cite{falke-etal-2019-ranking, kryscinski-etal-2020-evaluating, maynez-etal-2020-faithfulness, Fabbri2021, tang-etal-2023-understanding} or retrieval-augmented generation \cite{liu-etal-2023-evaluating, malaviya2024expertqa}, the entire claims are directly verified against the source documents. However, in cases where such documents are not readily available, such as in close-book generation, \citet{gao-etal-2023-rarr, min-etal-2023-factscore, Wang2023FactcheckGPTEF} decompose each generated sentence into atomic facts and then search for relevant documents to support each atomic fact. Alternatively, \citet{malaviya2024expertqa} directly search for relevant documents for each sentence as a whole.

There are two main approaches to verifying sentences against documents. The first involves training specialized fact-checkers specifically designed for factual consistency evaluation, which are primarily evaluated in the context of summarization \cite{kryscinski-etal-2020-evaluating, fabbri-etal-2022-qafacteval, goyal-durrett-2020-evaluating, laban-etal-2022-summac}. The second approach leverages LLMs as fact-checkers, particularly for evaluating LLM-generated responses from retrieval-augmented generation and closed-book generation \cite{min-etal-2023-factscore, Wang2023FactcheckGPTEF, malaviya2024expertqa, gao-etal-2023-rarr}. In this work, we bridge the gap between these two approaches by evaluating both specialized fact-checkers and LLM-based fact-checkers across all these settings using our new benchmark, \benchmarkname{}. We show that our best model can match GPT-4 performance and perform well in all settings without doing sentence decomposition.

\paragraph{Entailment Datasets} Our work contributes a new dataset for training textual entailment models over documents or document-chunks. Most prior entailment datasets have been human-authored \cite{bowman-etal-2015-large, williams-etal-2018-broad}, which is known to introduce artifacts \cite{gururangan-etal-2018-annotation}, or collected in the wild \cite{kamoi-etal-2023-wice}, which is challenging to scale. Past work has automatically generated contrast sets for NLI \cite{li-etal-2020-linguistically}. DocNLI \cite{yin-etal-2021-docnli} is a restructure of existing datasets where the length of most training examples cannot fit into the input limit of small models. Our work differs from these in its hand-built, synthetic nature to encourage multi-sentence and multi-fact reasoning, which is important to the task of fact-checking on grounding documents.

\section{Conclusion}

In this work, we introduce two  synthetic data generation methods that address key limitations of specialized fact-checkers by encouraging models to verify each atomic fact within a claim and reason across multiple sentences. We also present \benchmarkname{}, a new factual consistency evaluation dataset covering both closed-book and grounded generation settings. A model fine-tuned on our synthetic data outperforms all prior specialized fact-checkers on \benchmarkname{}, while being much cheaper than LLM-based fact-checkers.

\section*{Limitations}

\paragraph{Interpretation} Like many other specialized fact-checking models, our models do not reveal their internal decision-making processes, making it challenging to localize errors to particular mismatched spans of a claim or document. There are two ways to alleviate this issue. The first is to perform claim decomposition and check the models' prediction labels on each atomic fact, thereby localizing the error from the original claim. Our models show better performance compared to other specialized fact-checkers when using this claim decomposition method (Table~\ref{tab:decompose}), but this method can give greater interpretability. The second approach, which can be a future research direction, is to enable our best model, MiniCheck-\textsc{FT5}, or generative models in general, to provide reliable explanations in addition to the binary predictions. We believe a model can provide reliable explanations only if it can first correctly identify errors in our binary setting, which was the focus of this work. 

\paragraph{Multi-Document Reasoning} While our benchmark includes instances that evaluate models' ability to reason across multiple sentences, these datasets do not necessitate reasoning over evidence that is significantly separated or spread across various documents. Future research could focus on evaluating the performance of existing models in such scenarios, creating new labeled datasets of errors to expand our benchmark, and developing better fact-checking models to handle these challenges.

\paragraph{Synthetic Data} The effectiveness of our synthetic data is demonstrated by the improved performance when training on it across various model architectures. We find that using a pair of sentences to support a fact is a simple method that yields useful training data for our models. However, there are other possible strategies for how atomic facts or claims could be expanded into multiple sentences. We believe that constructing more complex and higher-quality data could be a future direction not only for this work but also for other related tasks. As LLMs continue to advance, the quality of the synthetic data generated using our approach is also expected to improve.

\paragraph{Language} Our models are trained exclusively on English data. Although the backbone model, Flan-T5, is trained on multilingual data, we have not systematically assessed how well our model's performance extends to other languages due to the absence of a human-annotated factual consistency evaluation dataset for LLM-generated outputs in non-English languages. We believe that developing a fact-checker that can perform well across multiple languages is important for future work.

\section*{Acknowledgments}

We would like to thank Jessy Li for comments on a draft of this work. This work was principally supported by a gift from Amazon as part of the UT-Amazon Science Hub. It was partially supported by NSF CAREER Award IIS-2145280, the NSF AI Institute for Foundations of Machine Learning (IFML), a grant from Open Philanthropy, a grant from the UT Austin Office of the Vice President for Research through the ``Creating Connections for National Security Research Grants'' program, and Good Systems,\footnote{https://goodsystems.utexas.edu/} a UT Austin Grand Challenge to develop responsible AI technologies.

\bibliography{anthology, custom}

\appendix

\section{Additional Analysis and Ablations}
\label{sec:analysis}

\subsection{Intrinsic Evaluation on C2D/D2C} \label{sec:intrinsic-eval}

Our model achieves strong overall performance, but we would like to have more insight as to whether it actually does well at the types of instances in D2C and C2D. We evaluate the performance of QAFactEval, SummaC-ZS, SummaC-CV, AlignScore, FT5-ANLI-L on our held-out synthetic data of C2D and D2C as a way to understand their ability to reason over multiple sentences and consider multiple atomic facts within a claim. There are 2K held-out instances from C2D and D2C, respectively, in the format of \texttt{(document, claim, label)} tuples. Performance is measured by BAcc.

\begin{table*}
\small
\centering
\renewcommand{\tabcolsep}{0.9mm}
\begin{tabular}{lcccccccccc|c}
\toprule
           & \multicolumn{10}{c}{\textbf{\benchmarkname{}} (\emph{without} threshold tuning)}                                                     \\
\cmidrule(r){2-11} 
\multirow{2}{*}{\begin{tabular}[c]{@{}l@{}}\textbf{Model}\\ \textbf{Name}\end{tabular}} &
  \multicolumn{2}{c}{\textbf{\textsc{AggreFact}}} &
  \multicolumn{2}{c}{\textbf{\textsc{TofuEval}}} &
  \multirow{2}{*}{\textbf{\textsc{Wice}}} &
  \multirow{2}{*}{\textbf{\textsc{Reveal}}} &
  \multirow{2}{*}{\begin{tabular}[c]{@{}c@{}}\textbf{\textsc{Claim}}\\ \textbf{\textsc{Verify}}\end{tabular}} &
  \multirow{2}{*}{\begin{tabular}[c]{@{}c@{}}\textbf{\textsc{Fact}}\\ \textbf{\textsc{Check}}\end{tabular}} &
  \multirow{2}{*}{\begin{tabular}[c]{@{}c@{}}\textbf{\textsc{Expert}}\\ \textbf{\textsc{QA}}\end{tabular}} &
   \multirow{2}{*}{\textbf{\textsc{Lfqa}}} &
  \multirow{2}{*}{\textbf{Avg}} \\
\cmidrule(r){2-3} \cmidrule(r){4-5} 

           & \textbf{CNN}  & \textbf{XSum} & \textbf{MediaS} & \textbf{MeetB} &      &      &      &      &      &     &  \\
\midrule
MiniCheck-\textsc{FT5}    & 69.9 & 74.3 & 73.6 & 77.3 & 72.2 & 86.2 & 74.6 & 74.7 & 59.0 & 85.2 & 74.7 \\
\midrule
\multicolumn{1}{r}{- \textsc{C2D}}  & 64.7 & 68.6 & 70.7 & 76.6 & 75.5 & 85.4 & 70.4 & 75.1 & 58.6  & 82.0 & 72.7 \color{red3}$\boldsymbol{(\downarrow 2.0)}$ \\
\multicolumn{1}{r}{- \textsc{D2C}}  & 62.9 & 70.6 & 70.1 & 75.9 & 74.0 & 83.1 & 67.1 & 75.5 & 58.0  & 77.9 & 71.5 \color{red3}$\boldsymbol{(\downarrow 3.2)}$ \\
\multicolumn{1}{r}{- \textsc{Both}}  & 54.7 & 59.4 & 54.1 & 55.6 & 61.5 & 77.1 & 57.4 & 65.0 & 51.9  & 63.0 & \,\,\,\,59.9 \color{red3}$\boldsymbol{(\downarrow 14.8)}$ \\
\bottomrule
\end{tabular}
\caption{\textbf{Ablation study on the training data.} Models are evaluated on the test set of \benchmarkname{} without threshold tuning. We show the average performance downgrade in red. -\textsc{Both} drops from 69.1 to 59.9 without threshold tuning.} \label{tab:ablation-study-full}
\end{table*}

\paragraph{Synthetic data ablations} Beside those specialized models, we fine-tune \texttt{flan-t5-large} on the training set of C2D (\textbf{FT5-C2D}) and D2C (\textbf{FT5-D2C}), respectively. We evaluate FT5-C2D on D2C as an out-of-distribution (OOD) evaluation set, and evaluate FT5-D2C on C2D.  

To demonstrate the necessity of steps in creating our synthetic data, we also create simplified versions of the synthetic data for both methods, denoted as \textsc{C2D-Simp} and \textsc{D2C-Simp}, each comprising 7K training examples, same as in C2D and D2C. For the \textsc{C2D-Simp} method, we ask GPT-4 to directly generate documents that support and not support a provided claim, with the requirement mentioned in the prompt that the inference on the claim should require reasoning over multiple sentences from a document. For the \textsc{D2C-Simp} method, we ask GPT-4 to generate summary sentences for Google News articles and come up with unsupported summaries by injecting errors into those summary sentences following the method in SummEdit \cite{laban-etal-2023-summedits}. We denote models trained on those simplified synthetic data as \textbf{FT5-C2D-S} and \textbf{FT5-D2C-S}. More details for creating these synthetic datasets can be found in Appendix~\ref{sec:baseline-methods}. 

\begin{figure}
\centering
\includegraphics[width=\linewidth, trim=0mm 0mm 00mm 0mm,clip]{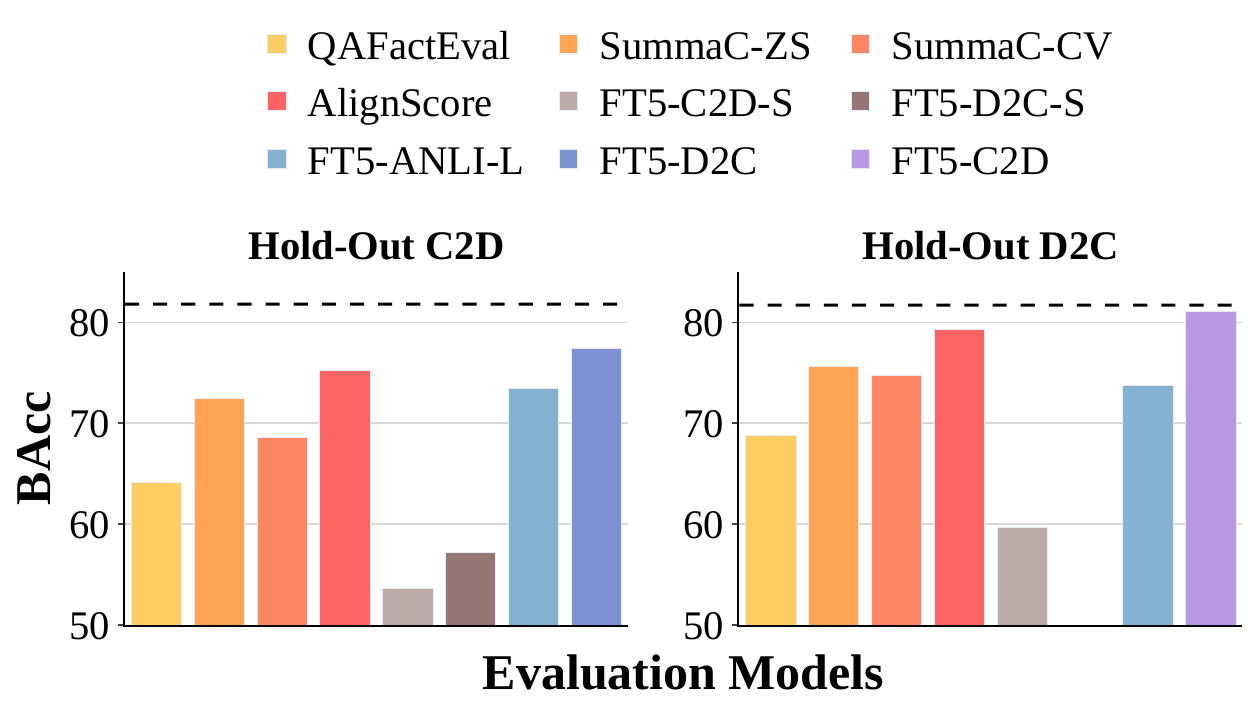}
\caption{Performance of fact-checkers on the held-outs sets of C2D (left) and D2C (right). The black dashed line shows the in-distribution performance of FT5-C2D (left) and FT5-D2C (right).} \label{fig:eval_on_synthetic}
\end{figure}

\begin{table*}[t]
\small
\centering
\renewcommand{\tabcolsep}{1.5mm}
\begin{tabular}{lcccccccccc|c}
\toprule
           & \multicolumn{10}{c}{\textbf{\benchmarkname{}} (\emph{with} threshold tuning)}                                                     \\
\cmidrule(r){2-11} 
\multirow{2}{*}{\begin{tabular}[c]{@{}l@{}}\textbf{Model}\\ \textbf{Name}\end{tabular}} &
  \multicolumn{2}{c}{\textbf{\textsc{AggreFact}}} &
  \multicolumn{2}{c}{\textbf{\textsc{TofuEval}}} &
  \multirow{2}{*}{\textbf{\textsc{Wice}}} &
  \multirow{2}{*}{\textbf{\textsc{Reveal}}} &
  \multirow{2}{*}{\begin{tabular}[c]{@{}c@{}}\textbf{\textsc{Claim}}\\ \textbf{\textsc{Verify}}\end{tabular}} &
  \multirow{2}{*}{\begin{tabular}[c]{@{}c@{}}\textbf{\textsc{Fact}}\\ \textbf{\textsc{Check}}\end{tabular}} &
  \multirow{2}{*}{\begin{tabular}[c]{@{}c@{}}\textbf{\textsc{Expert}}\\ \textbf{\textsc{QA}}\end{tabular}} &
  \multirow{2}{*}{\textbf{\textsc{Lfqa}}} &
  \multirow{2}{*}{\textbf{Avg}} \\
\cmidrule(r){2-3} \cmidrule(r){4-5} 

           & \textbf{CNN}  & \textbf{XSum} & \textbf{MediaS} & \textbf{MeetB} &      &      &      &      &      &    &   \\
\midrule
T5-NLI-Mixed & 59.9 & 56.1 & 60.6 & 55.9 & 58.0 & \cellcolor{green2}87.6 & 61.5 & 69.3 & 56.8 & 62.8 & 62.9 \\
DAE & 58.6 & 67.1 & 67.6 & 63.2 & 57.1 & 83.8 & 71.1 & 72.6 & \cellcolor{green2}58.6 & 68.5 & 68.5 \\
QAFactEval & 63.9 & 63.7 & 64.0 & 66.8 & 65.8 & 85.3 & 73.3 & 72.1 & 57.6  & 81.5 & 69.4 \\
SummaC-ZS   & 63.0 & 67.2 & \cellcolor{green2}69.5 & 70.0 & 61.8 & \cellcolor{green2}86.4 & 69.9 & \cellcolor{green2}75.7 & 57.5  & 82.0 & 70.4 \\
SummaC-CV   & \cellcolor{green2}67.6 & 69.7 & 68.2 & 71.0 & 66.1 & \cellcolor{green2}87.3 & 71.3 & 74.3 & \cellcolor{green2}59.3  & 71.3 & 71.3 \\
AlignScore & 62.6 & 69.6 & \cellcolor{green2}71.6 & 71.8 & 66.8 & 85.3 & 72.9 & \cellcolor{green2}76.5 & \cellcolor{green2}59.2  & \cellcolor{green2}85.6 & 72.2 \\
% FT5-ANLI-L & 64.5 & 70.9 & \cellcolor{green2}70.3 & \cellcolor{green2}75.0 & \cellcolor{green2}73.4 & 84.8 & 67.7 & \cellcolor{green2}75.4 & 57.5 & 71.1 \\
\midrule
MiniCheck-\textsc{Rbta}  & 64.6 & 70.2 & \cellcolor{green2}71.1 & \cellcolor{green2}75.2 & \cellcolor{green2}73.7 & \cellcolor{green2}88.0 & \cellcolor{green2}77.1 & \cellcolor{green2}77.3 & \cellcolor{green2}58.4  & \cellcolor{green2}84.2 & 73.7 \\
MiniCheck-\textsc{Dbta}  & 63.4 & \cellcolor{green2}74.7 & 69.1 & \cellcolor{green2}72.8 & \cellcolor{green2}76.3 & \cellcolor{green2}87.4 & \cellcolor{green2}75.5 & \cellcolor{green2}76.0 & \cellcolor{green2}58.8  & \cellcolor{green2}84.1 & 73.8 \\
MiniCheck-\textsc{FT5}   & \cellcolor{green2}71.5 & \cellcolor{green2}74.8 & \cellcolor{green2}73.7 & \cellcolor{green2}76.7 & \cellcolor{green2}75.0 & \cellcolor{green2}86.4 & \cellcolor{green2}73.8 & \cellcolor{green2}76.4 & \cellcolor{green2}58.6  & \cellcolor{green2}84.4 & \cellcolor{green2}75.1 \\
\bottomrule
\end{tabular}
\caption{\textbf{Performance of models on the test set of \benchmarkname{} \emph{with} threshold tuning on the validation set.} Balanced accuracy is computed for each model on the 10 datasets in \benchmarkname{}, and the average is computed. In each dataset, a factuality metric selects a threshold based on the performance on the corresponding validation set.} \label{tab:result-threshold-tuning}
\end{table*}

\paragraph{Synthetic data needs to be carefully constructed for an fact-checker to work well.} Figure~\ref{fig:eval_on_synthetic} shows the in-distribution performance of FT5-C2D and FT5-D2C as optimal performance in each sub-figure, represented by the black dashed lines. We observe that models trained on synthetic data with simplified construction steps (\textsc{C2D-Simp} and \textsc{D2C-Simp}) fail to develop the desired properties we expect. FT5-D2C-S performs even worse than random chance on the held-out set of C2D. In contrast, models trained on C2D and D2C outperform all other fact-checkers on the OOD held-out set of D2C and C2D, respectively. Additionally, we note that the model trained on 163K ANLI data points fail to reach the performance of models trained solely on 7K synthetic data. Our synthetic data generation methods can effectively encourage models to pay more attention to multiple atomic facts and reason over multiple sentences, even with a limited amount of training data.

\begin{table*}[t]
\small
\centering
\renewcommand{\tabcolsep}{1mm}
\begin{tabular}{lcccccccccc|c}
\toprule
           & \multicolumn{10}{c}{\textbf{\benchmarkname{}} (decomposition - \emph{without} threshold tuning)}                                                     \\
\cmidrule(r){2-11} 
\multirow{2}{*}{\begin{tabular}[c]{@{}l@{}}\textbf{Model}\\ \textbf{Name}\end{tabular}} &
  \multicolumn{2}{c}{\textbf{\textsc{AggreFact}}} &
  \multicolumn{2}{c}{\textbf{\textsc{TofuEval}}} &
  \multirow{2}{*}{\textbf{\textsc{Wice}}} &
  \multirow{2}{*}{\textbf{\textsc{Reveal}}} &
  \multirow{2}{*}{\begin{tabular}[c]{@{}c@{}}\textbf{\textsc{Claim}}\\ \textbf{\textsc{Verify}}\end{tabular}} &
  \multirow{2}{*}{\begin{tabular}[c]{@{}c@{}}\textbf{\textsc{Fact}}\\ \textbf{\textsc{Check}}\end{tabular}} &
  \multirow{2}{*}{\begin{tabular}[c]{@{}c@{}}\textbf{\textsc{Expert}}\\ \textbf{\textsc{QA}}\end{tabular}} &
  \multirow{2}{*}{\textbf{\textsc{Lfqa}}} &
  \multirow{2}{*}{\textbf{Avg}} \\
\cmidrule(r){2-3} \cmidrule(r){4-5} 

           & \textbf{CNN}  & \textbf{XSum} & \textbf{MediaS} & \textbf{MeetB} &      &      &      &      &      &      \\
\midrule
GPT-4   & 70.1 & 74.3 & 70.9 & 79.4 & 77.6 & 87.4 & 73.2 & 79.9 & 59.6  & 84.9 & 75.6  \color{greenarrow}$(\boldsymbol{\uparrow 0.3})$\\
\midrule
SummaC-CV   & 65.2 & 51.2 & 58.0 & 55.2 & 50.9 & 66.2  & 69.6  & 53.4 & 53.5  & 65.0 & 58.8 \color{red3}$(\boldsymbol{\downarrow 3.3})$ \\
QAFactEval & 61.7 & 60.4 & 65.0 & 60.2 & 59.1  & 81.7  & 68.6  & 66.1 & 54.6   & 73.0 & 64.6 \color{red3}$(\boldsymbol{\downarrow 1.9})$ \\
SummaC-ZS   & 62.4 & 64.5 & 64.5 & 70.0 & 64.8 & 85.6  & 70.0  & 75.2 & 56.3   & 77.2 & 69.1 \color{greenarrow}$(\boldsymbol{\uparrow 1.2})$\\
AlignScore & 65.6 & 68.3 & 71.2 & 73.2 & 63.9 & 86.2 & 73.7 & 74.3 & 57.6  & 82.6 & 71.5 \color{greenarrow}$(\boldsymbol{\uparrow 1.1})$\\
\midrule
MiniCheck-\textsc{Rbta} & 65.0 & 72.1 & 72.6 & 75.1 & 66.9 & 88.5 & 76.1 & 73.3 & 58.0  & 84.3 & 73.2 \color{greenarrow}$(\boldsymbol{\uparrow 0.5}$)\\
MiniCheck-\textsc{Dbta}  & 59.9 & 73.3 & 67.9 & 74.5 & 75.0 & 88.2 & 72.5 & 73.0 & 57.9 & 84.5 & 72.7 \color{greenarrow}$(\boldsymbol{\uparrow 0.1})$ \\
MiniCheck-\textsc{FT5}    & 65.9 & 71.7 & 69.2 & 77.4 & 72.9 & 87.0 & 73.5 & 74.7 & 57.5 & 83.2 & 73.3 \color{red3}$(\boldsymbol{\downarrow 1.4})$\\
\bottomrule
\end{tabular}
\caption{Performance of models on the test set of \benchmarkname{} by aggregating predictions on decomposed claims. We include the performance change compared to predicting using original claims from Table~\ref{tab:result-zs} (red for worse performance and green for better performance).} \label{tab:decompose-full}
\end{table*}

\begin{table*}
\small
\centering
\renewcommand{\tabcolsep}{1mm}
\begin{tabular}{lcccccccccc|c}
\toprule
           & \multicolumn{10}{c}{\textbf{\benchmarkname{}} (decontextualization - \emph{without} threshold tuning)}                                                     \\
\cmidrule(r){2-11} 
\multirow{2}{*}{\begin{tabular}[c]{@{}l@{}}\textbf{Model}\\ \textbf{Name}\end{tabular}} &
  \multicolumn{2}{c}{\textbf{\textsc{AggreFact}}} &
  \multicolumn{2}{c}{\textbf{\textsc{TofuEval}}} &
  \multirow{2}{*}{\textbf{\textsc{Wice}}} &
  \multirow{2}{*}{\textbf{\textsc{Reveal}}} &
  \multirow{2}{*}{\begin{tabular}[c]{@{}c@{}}\textbf{\textsc{Claim}}\\ \textbf{\textsc{Verify}}\end{tabular}} &
  \multirow{2}{*}{\begin{tabular}[c]{@{}c@{}}\textbf{\textsc{Fact}}\\ \textbf{\textsc{Check}}\end{tabular}} &
  \multirow{2}{*}{\begin{tabular}[c]{@{}c@{}}\textbf{\textsc{Expert}}\\ \textbf{\textsc{QA}}\end{tabular}} &
  \multirow{2}{*}{\textbf{\textsc{Lfqa}}} &
  \multirow{2}{*}{\textbf{Avg}} \\
\cmidrule(r){2-3} \cmidrule(r){4-5} 

           & \textbf{CNN}  & \textbf{XSum} & \textbf{MediaS} & \textbf{MeetB} &      &      &      &      &      &     &  \\
\midrule
GPT-4   & 66.7 & 76.5 & 71.5 & 79.2 & 80.3 & 87.0 & 66.2 & 79.9 & 60.1  & 86.4 & 75.3  \color{black}$(+ \boldsymbol{0.0})$\\
\midrule
SummaC-CV   & 65.2 & 54.5 & 62.6 & 62.1 & 52.5 & 67.8  & 69.0  & 53.4 & 54.7   & 66.5 & 60.8 \color{red3}$(\boldsymbol{\downarrow 1.3})$ \\
QAFactEval & 54.3 & 62.1 & 62.8 & 64.5 & 62.7  & 82.8  & 71.7  & 66.9 & 56.5   & 80.3 & 66.4 \color{red3}$(\boldsymbol{\downarrow 0.1})$ \\
SummaC-ZS   & 51.1 & 61.5 & 67.5 & 71.4 & 63.0 & 85.8  & 69.0  & 75.2 & 56.4   & 76.1 & 67.7 \color{red3}$(\boldsymbol{\downarrow 0.2})$\\
AlignScore & 52.4 & 71.4 & 68.8 & 72.2 & 65.1 & 85.5 & 69.1 & 74.3 & 59.6  & 85.2 & 70.4 \color{black}$(+ \boldsymbol{0.0})$\\
\midrule
MiniCheck-\textsc{Rbta} & 63.7 & 70.8 & 70.9 & 75.6 & 64.3 & 88.9 & 76.0 & 73.3 & 57.5  & 83.2& 72.4 \color{red3}$(\boldsymbol{\downarrow 0.3})$\\
MiniCheck-\textsc{Dbta}  & 64.2 & 71.0 & 69.6 & 69.1 & 63.3 & 87.5 & 73.6 & 73.0 & 57.8 &  83.2& 71.2 \color{red3}$(\boldsymbol{\downarrow 1.4})$ \\
MiniCheck-\textsc{FT5}    & 69.9 & 74.3 & 74.0 & 75.6 & 69.7 & 86.2 & 73.0 & 74.7 & 58.4 & 85.2 & 74.1 \color{red3}$(\boldsymbol{\downarrow 0.6})$\\
\bottomrule
\end{tabular}
\caption{Performance of models on the test set of \benchmarkname{} by doing claim decontextualization where it is applicable. We include the performance change compared to predicting using original claims from Table~\ref{tab:result-zs}.} \label{tab:dectx-full}
\end{table*}

\subsection{Ablation of D2C/C2D} 

We observe that there are still gaps between the optimal performance and that achieved by other specialized fact-checkers in Figure~\ref{fig:eval_on_synthetic}. Notably, AlignScore demonstrates the best performance among the four specialized metrics, but its performance can still be outperformed by FT5-C2D and FT5-D2C. As shown in Table~\ref{tab:result-zs}, we can improve AlignScore's performance on the benchmark by fine-tuning it on this small amount of data, effectively equipping it with the desired properties.

To further investigate, we conducted an ablation study on our top-performing model, MiniCheck-\textsc{FT5}, by removing our synthetic data from the training set. The results, presented in Table~\ref{tab:ablation-study-full}, reveal that the two types of synthetic data complement each other. Notably, the model performs poorly when trained solely on the ANLI subset, with a performance drop of nearly 10\% in the absence of threshold tuning. However, the addition of either 7K C2D or 7K D2C to the training data significantly enhances the model's performance and robustness.

\subsection{Human Evaluation of Synthetic Data} \label{sec:human-eval}

While we use automatic entailment checks to ensure label quality in the training data construction pipeline, we conduct a small scale human evaluation to measure the quality of the generated data.

We randomly chose 40 (document, claim) pairs from each of the C2D and D2C training data. Three authors of the work independently annotated these 40 $\times$ 2 datapoints as supported or not supported (without seeing the gold label), with an average annotation time of around 1 min and 40 seconds per example. Among the three annotators, we computed an annotation agreement using Fleiss’ Kappa, obtaining a score of 0.51 for the C2D samples and 0.70 on the D2C samples, indicating moderate to substantial agreement. The annotators adjudicated cases with disagreement to reach a consensus on the final factuality of the labels. We refer to these as the ground truth labels.

Compared to these ground truth labels, the labels given to our training samples have an accuracy of 80\% on C2D and 78\% on D2C, respectively. We calculated similar accuracy values for the human annotators, and the average across the three annotators’ accuracies was 85\% on C2D and 88\% on D2C. These accuracies demonstrate that the automatic labels are a bit lower-quality than single-annotator labels on D2C, but close on C2D.

Many of the disagreement cases are classic cases of subjectivity in NLI \cite{pavlick-kwiatkowski-2019-inherent, chen-etal-2020-uncertain, nie-etal-2020-learn}. For example, for the C2D claim ``\emph{Located in the Scottish Highlands, she lived with and later married James Ballard in Spean Bridge.}'', the generated passage only references the couple living in the town of Spean Bridge after the marriage, making it unclear whether or not they got married there. However, this is a reasonable supposition to make. Our results show that our data is useful for training in spite of these subjective examples.

\section{Additional Results}

\subsection{Results using Threshold Tuning} \label{sec:result-tuning-thresh}

As shown in Table~\ref{tab:result-threshold-tuning}, AlignScore achieves the highest overall performance (72.2\%) on \benchmarkname{} among fact-checkers from prior work. However, by fine-tuning AlignScore's backbone RoBERTa model on our 14K synthetic data (MiniCheck-\textsc{Rbta}), we surpass AlignScore's performance by 1.5\%. This improvement is more significant in the setting without threshold tuning (Section~\ref{sec:results}). Remarkably, this boost in performance is attained using a dataset that constitutes less than 0.3\% of the total data on which AlignScore was initially trained (Table~\ref{tab:model-card}). This finding highlights the potential of curated synthetic data in enhancing the performance of state-of-the-art fact-checkers. Overall, our models achieve new state-of-the-art among specialized models under the threshold tuning setting.

\paragraph{Our synthetic data enhances model robustness and performance in the absence of threshold tuning.} Comparing Table~\ref{tab:result-zs} with Table~\ref{tab:result-threshold-tuning}, we see that the performance of specialized fact-checkers decreases without threshold tuning. However, \emph{MiniCheck-\textsc{\emph{FT5}} only drops by 0.4\%} compared to larger drops for other systems, such as 9.2\% for SummaC-CV. These results suggest that our synthetic data not only improves overall performance but also enhances the robustness of models across the domains of our benchmark, enabling them to maintain strong performance even without threshold tuning.

\subsection{Full Results Using Claim Decomposition}

In Table~\ref{tab:decompose-full}, we show the performance (and performance changes) of GPT-4 and a subset of specialized fact-checkers across all datasets from \benchmarkname{}, using claim decomposition to determine the factuality label for each claim.

\subsection{Full Results Using Claim Decontextualization}

In Table~\ref{tab:dectx-full}, we show the performance (and performance changes) of GPT-4 and a subset of specialized fact-checkers across all datasets from \benchmarkname{}, using claim decontextualization when applicable. In particular, we perform claim decontextualization on \textsc{TofuEval}-MediaS, \textsc{TofuEval}-MeetB, \textsc{Wice}, \textsc{Reveal}, \textsc{ClaimVerify}, \textsc{ExpertQA}, and \textsc{Lfqa}. Note that the claim decontextualization step add a non-negligible amount of cost, as shown in Table~\ref{tab:llm-throughput}.

\subsection{Results Predicting All Errors}

When a grounding document is relevant to multiple sentences in a response, it becomes feasible to ask an LLM-based fact-checker to simultaneously predict the factuality labels for all sentences, thereby reducing cost. We investigate this approach using GPT-4 on the \textsc{TofuEval} datasets, where we provide GPT-4 with a document and the entire summary, asking the model to predict the factuality labels for all summary sentences at once, denoted as \textbf{GPT4-Full}. The prompt is shown in Table~\ref{tab:summ-prompt}. Table~\ref{tab:result-use-summ} shows that GPT4-Full achieves performance similar to predicting the factual label for each summary sentence individually. The inference cost on the \textsc{TofuEval} test set can be reduced from \$16.7 to \$6.72 with this method.

However, in retrieve-then-generate and post-hoc grounding settings, evidence is typically retrieved for each claim separately, meaning no document can be shared across claims in a single response, and thus the inference cost is barely reduced.

\begin{table}
\centering
\begin{tabular}{lcc}
\toprule
\multirow{2}{*}{\begin{tabular}[c]{@{}l@{}}\textbf{Model}\\ \textbf{Name}\end{tabular}} & \multicolumn{2}{c}{\textbf{\textsc{TofuEval}}} \\
\cmidrule{2-3}
       & \textbf{MediaS} & \textbf{MeetB} \\
\midrule
GPT-4  &  71.4    &   79.9    \\
GPT-4-Full &   72.3   &   79.7   \\
\bottomrule
\end{tabular} 
\caption{Comparison of GPT-4 and GPT-4-Full on \textsc{TofuEval}, a dataset where sentences from an LLM-generated response share the same grounding document.}  \label{tab:result-use-summ}
\end{table}

\section{\benchmarkname{} Details} \label{sec:benchmark-stats}

\subsection{Dataset Descriptions}

\paragraph{\textsc{AggreFact}} \cite{tang-etal-2023-understanding} is a factual consistency evaluation benchmark for new summarization, targeting \textbf{CNN(/DM)} \cite{Nallapati2016} and \textbf{XSum} \cite{narayan-etal-2018-dont}. Our focus is on the SOTA sets within \textsc{AggreFact}, where summaries are generated from SOTA fine-tuned summarizers, since their analysis suggests that summaries are more challenging to evaluate for factual consistency compared to summaries generated by pre-SOTA summarizers. Data in \textsc{AggreFact} comes from 9 factual consistency evaluation datasets on CNN or XSum, including widely used ones such as SummaC \cite{laban-etal-2022-summac}, FRANK \cite{pagnoni-etal-2021-understanding}, and SummEval \cite{Fabbri2021}. Check Appendix~\ref{sec:benchmark-stats} for a complete set of evaluation datasets in \textsc{AggreFact}. Because CNN/DM and XSum feature quite different styles of summaries, we report these numbers separately in our benchmark. However, we do not otherwise report results on the smaller datasets within the \textsc{AggreFact} SOTA subset.

\paragraph{\textsc{TofuEval}} \cite{tang2024tofueval} is a factual consistency evaluation benchmark for dialogue summarization, targeting \textbf{MediaS}um (interviews, \citet{zhu-etal-2021-mediasum}) and \textbf{Meet}ing\textbf{B}ank (city council meetings, \citet{hu-etal-2023-meetingbank}). It includes topic-focused dialogue summaries generated by 6 LLMs, with sentence-level factual consistency annotations by linguists. 

\paragraph{\textsc{Wice}} \cite{kamoi-etal-2023-wice} is a textual entailment dataset that consists of naturally occurring claims from Wikipedia and their cited documents. Based on its cited documents, each claim is labeled as \texttt{supported}, \texttt{partially-supported}, or \texttt{non-supported}.

\paragraph{\textsc{Reveal}} \cite{jacovi2024chainofthought} is a benchmark dataset that evaluates the correctness of reasoning chains generated by LLMs in the context of open-domain question-answering. The dataset includes annotations at the sentence level, covering various aspects of response correctness. For our dataset, we focus on the subset of sentences that have \emph{attribution} annotations, which indicate whether a sentence in a reasoning chain can be attributed to information retrieved from Wikipedia paragraphs with three label categories:  \texttt{fully attributable}, \texttt{partially attributable}, or \texttt{contradictory}.

\paragraph{\textsc{ClaimVerify}} \cite{liu-etal-2023-evaluating} evaluates the correctness of responses from four generative search engines in answering user queries. Similar to \textsc{Wice}, the dataset contains annotations on whether check-worthy sentences from the engines' responses can be fully supported by their associated cited documents. The dataset contains binary-level factual consistency annotations for each cite-worthy sentence.

\paragraph{\textsc{FactCheck}-GPT} \cite{Wang2023FactcheckGPTEF} contains factual consistency annotations for LLMs' responses to search queries. In this dataset, each sentence from LLMs' responses is first decomposed into atomic facts and those atomic facts are then further decontextualized so that they can stand alone. Finally, each worth-checking and decontextualized atomic fact is labeled as \texttt{completely support}, \texttt{partially support}, \texttt{refute}, or \texttt{irrelevant}. We include those decontextualized atomic facts and their corresponding documents in the benchmark.

\paragraph{\textsc{ExpertQA}} \cite{malaviya2024expertqa} contains responses from 6 different systems to queries curated by experts from 32 fields. These systems answer queries either in a close-book fashion with/without in-line citations, or based on retrieved document(s). For each sentence in the response, the sentence is verified against the concatenation of cited or retrieved document(s), if any. We include examples where documents are judged as \texttt{complete}, \texttt{partial}, or \texttt{incomplete} in supporting the corresponding sentences. We do not include human edited claims and evidence in our benchmark. 

\paragraph{\textsc{Lfqa}} \cite{chen2023understanding} contains LLM-generated responses to questions from the ELI5 (``Explain Like I’m Five'') dataset \cite{fan-etal-2019-eli5}. LLMs generate responses based on documents that are either retrieved by humans, models, or randomly selected. Human annotators then evaluate each sentence in the LLM-generated responses against the corresponding document set, classifying them into \texttt{supported}, \texttt{partially supported}, or \texttt{not supported}.

\begin{table}
\small
\centering
\renewcommand{\tabcolsep}{1.2mm}
\begin{tabular}{ccccccc}
\toprule
\multicolumn{2}{c}{\textbf{Dataset}} &
 \textbf{Split} &
  \textbf{Size} &
  \begin{tabular}[c]{@{}c@{}}\textbf{Doc}\\ \textbf{Len}\end{tabular} &
  \begin{tabular}[c]{@{}c@{}}\textbf{Claim}\\ \textbf{Len}\end{tabular} &
  \begin{tabular}[c]{@{}c@{}} \textbf{\% of}\\  \textbf{Neg}\end{tabular} \\
\midrule
\multirow{4}{*}{\textsc{AggreFact}}            & \multirow{2}{*}{CNN}                      & dev  & 459  & 558  & 56 & 13\%\\
                                              &                                           & test & 558  & 563  & 58 & 10\% \\
                                              \cmidrule(r){2-7} 
                                              & \multirow{2}{*}{XSum}                     & dev  & 777  & 374  & 26 & 49\% \\
                                              &                                           & test & 558  & 370  & 25 & 49\%  \\
\midrule
\multirow{4}{*}{\textsc{TofuEval}} &
  \multirow{2}{*}{Media} &
  dev &
  1800 &
  990 &
  21 & 20\%\\
                                              &                                           & test & 726  & 922  & 21 &  24\%\\
                                               \cmidrule(r){2-7} 
                                              & \multirow{2}{*}{MeetB}                    & dev  & 1607 & 930  & 22 & 18\%\\
                                              &                                           & test & 772  & 915  & 22 & 19\% \\
\midrule
\multicolumn{2}{c}{\multirow{2}{*}{\textsc{Wice}}}                                        & dev  & 349  & 1622 & 27 & 67\%\\
\multicolumn{2}{c}{}                                                                      & test & 358  & 1683 & 28 & 69\%\\
\midrule
\multicolumn{2}{c}{\multirow{2}{*}{\textsc{Reveal}}}                                      & dev  & 1656 & 104  & 11 & 77\%\\
\multicolumn{2}{c}{}                                                                      & test & 1710 & 103  & 11 & 77\%\\
\midrule
\multicolumn{2}{c}{\multirow{2}{*}{\begin{tabular}[c]{@{}c@{}}\textsc{Claim}\\ \textsc{Verify}\end{tabular}}} &
  dev &
  1093 &
  1874 &
  21 & 25\%\\
\multicolumn{2}{c}{}                                                                      & test & 1088 & 1841 & 22 & 27\%\\
\midrule
\multicolumn{2}{c}{\multirow{2}{*}{\begin{tabular}[c]{@{}c@{}}\textsc{Fact}\\ \textsc{Check}\end{tabular}}} & dev  & 1537 & 100  & 14 & 83\%\\
\multicolumn{2}{c}{}                                                                      & test & 1566 & 100  & 13 & 76\%\\
\midrule
\multicolumn{2}{c}{\multirow{2}{*}{\begin{tabular}[c]{@{}c@{}}\textsc{Expert}\\ \textsc{QA}\end{tabular}}}  & dev  & 3773 & 491  & 29 & 22\%\\
\multicolumn{2}{c}{}                                                                      & test & 3702 & 506  & 29 & 20\%\\
\midrule
\multicolumn{2}{c}{\multirow{2}{*}{\textsc{Lfqa}}}                                        & dev  &  2029 & 383 & 25 & 43\%\\
\multicolumn{2}{c}{}                                                                      & test &  1911 & 380 & 25 & 41\%\\
\bottomrule
\end{tabular}
\caption{\textbf{Statistics of datasets in \benchmarkname{}.} We show the size of datasets, the average length of documents and claims, and the proportion of unsupported claims.} \label{tab:benchmark-stats}
\end{table}

\subsection{Label Unification} 

For \textsc{AggreFact},  \textsc{TofuEval}, and  \textsc{ClaimVerify}, we keep using the binary label from the original work. For the remaining datasets, we map \texttt{supported}, \texttt{fully attributable},  \texttt{completely support}, and \texttt{complete} to \texttt{supported}, and \texttt{unsupported} otherwise.

\subsection{Excluded Datasets}

We excluded \textbf{\textsc{HaluEval}} \cite{li-etal-2023-halueval} and \textbf{\textsc{SummEdits}} \cite{laban-etal-2023-summedits} from our benchmark since they are synthetic, with errors in summaries generated via instruction prompts that guide the model to intentionally make errors in summaries. These errors are unnatural and do not fit with our goal of detecting true LLM generation errors. \textbf{\textsc{FActScore}} \cite{min-etal-2023-factscore} contains naturally generated biographies from LLMs and has human-annotated labels of individual atomic facts. However, humans could potentially search different articles to verify the correctness of those sentences than the ones that models retrieve. As a result, a non-negligible fraction of the claims in the dataset appear mislabeled from the standpoint of the fact-checking on grounded documents task.

\subsection{Statistics}

The statistics of \benchmarkname{} can be found in Table~\ref{tab:benchmark-stats}. Our use of these datasets is for research purposes only, which is consistent with their intended use.

\textsc{AggreFact} contains the following 9 factual consistency evaluation datasets on CNN or XSum: FactCC \cite{kryscinski-etal-2020-evaluating}, Wang'20 \cite{wang-etal-2020-asking},  SummEval \cite{Fabbri2021}, Polytope \cite{huang-etal-2020-achieved}, Cao'22 \cite{cao-etal-2022-hallucinated}, XSumFaith \cite{maynez-etal-2020-faithfulness}, FRANK \cite{pagnoni-etal-2021-understanding}, Goyal'21 \cite{goyal-durrett-2021-annotating}, and CLIFF \cite{cao-wang-2021-cliff}.

\section{Synthetic Data Details} \label{sec:synthetic-data-stats}

\paragraph{Source of Data} In our C2D method, we choose around 400 claims from Wikipedia that have cited web articles \cite{kamoi-etal-2023-wice, Petroni2023} to generate synthetic documents. In our D2C method, we scraped around 300 Google News articles since November 2023 from diverse topic categories, including science, politics, world, entertainment, business, and technology. Each document is approximately 500 words. Statistics of our generated data can be found in Table~\ref{tab:train-data-stats} and~\ref{tab:length-distribution}. See Appendix~\ref{sec:quality-check} for how we maintain the quality of our synthetic data.

\begin{table}
\centering
\renewcommand{\tabcolsep}{1.6mm}
\begin{tabular}{lcccccc}
\toprule
                     & \textbf{Label} & \textbf{Min.} & \textbf{25\%} & \textbf{50\%} & \textbf{75\%} & \textbf{Max.} \\
\midrule
\multirow{2}{*}{C2D} & 1     & 72  & 147  & 182  & 242  & 359 \\
                     & 0     & 66  & 136  & 171  & 234  & 359 \\
\midrule
\multirow{2}{*}{D2C} & 1     & 85  & 139  & 162  & 186  & 427 \\
                     & 0     & 84  & 138  & 161  & 186  & 493 \\
\bottomrule
\end{tabular}
\caption{Length distribution of generated documents. We use the NLTK package for word tokenizatiopn.} \label{tab:length-distribution}
\end{table}

\paragraph{Characteristics of Synthetic Data} It is worth noting that constructing our synthetic dataset involves using human-written or naturally generated claims, which sets it apart from prior synthetic data generation methods used to train fact-checkers for text summarization. These methods, such as entity swapping and sentence negation \cite{kryscinski-etal-2020-evaluating, goyal-durrett-2021-annotating}, were designed to target specific error types that occurred in claims from earlier summarization models. However, as errors from generative models progress \cite{tang-etal-2023-understanding} and new error types emerge from LLMs, focusing on specific error types may not generalize well to unseen datasets with potentially novel errors.

Examples of synthetic data for C2D and D2C can be found in Table~\ref{tab:C2D-example} and~\ref{tab:D2C-example}.

\paragraph{Data Rejection Rate} Since we use GPT-4 for filtering out low-quality examples in our C2D method, we report the rejection rate at different steps that require entailment checks.

During the \emph{atomic fact expansion} step (step 2), 6\% of the final generated sentence pairs, when combined, could not support the original atomic fact. In the \emph{supporting document generation} step (step 3), 5\% of the final documents failed the entailment check. For the \emph{non-supporting document generation} step (step 4), 53\% of the final documents still supported the claim, and these documents were not included in our constructed data. These filtering steps are crucial for improving the training dataset's quality.

\section{Fact-Checking Model Details}

\subsection{LLM-Based Fact-Checkers} \label{sec:llm-info}

We use the official APIs for LLM-based fact-checkers. The checkpoints we use for LLMs can be found in Table~\ref{tab:llm-checkpoint}. The inference prompt is the same for all LLMs and can be found in Table~\ref{tab:zero-shot-eval}. We use a temperature of zero to collect deterministic outputs, which is typical from previous work.

\begin{table}
\centering
\begin{tabular}{lc}
\toprule
\textbf{Model}         & \textbf{Checkpoint}             \\
\midrule
Gemini-Pro    &  \texttt{gemini-1.0-pro}       \\
PaLM2-Bison   &  \texttt{chat-bison@001}        \\
Mistral-8x7B  & \texttt{open-mixtral-8x7b}      \\
Mistral-Large & \texttt{mistral-large-2402}     \\
Claude-2.1    & \texttt{claude-2.1}             \\
Claude-3 Opus & \texttt{claude-3-opus-20240229} \\
GPT-3.5       & \texttt{gpt-3.5-turbo-0125}     \\
GPT-4         & \texttt{gpt-4-0125-preview}    \\
\bottomrule
\end{tabular}
\caption{LLM checkpoints} \label{tab:llm-checkpoint}
\end{table}

\subsection{Specialized Fact-Checkers} \label{sec:evaluators-detail}

\paragraph{QAFactEval} \cite{fabbri-etal-2022-qafacteval} is a QA-based fact-checker with optimized components for answer selection, question answering, question generation, and answer overlap calculation. It selects spans as answers from a summary sentence, generates questions based on these answers, and then answers these questions using the source document. Finally, it computes an overall overlap score for the summary sentence by comparing the selected spans from the summary sentence with the answers derived from the source document, given the generated questions. QAFactEval produces scores on a continuous scale ranging from 0 to 5. In our experiments, we use the default model and hyperparameters as provided by the authors.

\paragraph{DAE} \cite{goyal-durrett-2020-evaluating,goyal-durrett-2021-annotating} is an entailment-based fact-checker that evaluates the factual consistency of each dependency arc in a summary sentence. It independently verifies whether the semantic relationship of each dependency arc is factually supported by the source document. Finally, it aggregates the scores for all dependency arcs to compute an overall sentence-level factuality score ranging from 0 to 1. In our experiments, we use the default model and hyperparameters as provided by the authors in \citet{goyal-durrett-2021-annotating}.

\paragraph{SummaC-ZS} \cite{laban-etal-2022-summac} is an entailment-based fact-checker. To evaluate a summary sentence $c_i$, it divides the source document $D_i$ into a set of sentences or paragraphs $D_i = \{d_{i,1}, \ldots, d_{i,|d|}\}$, and the score for $c_i$ is determined by the highest score among all $(d_{i,j}, c_i)$ pairs, i.e., $\mathrm{score}(c_i) = \mathrm{max}_j M(d{i,j}, c_i)$. For a multi-sentence summary, the final score is calculated as the average of the individual sentence scores. In our experiments, we do not use the authors' default setting of splitting the document $D_i$ into sentences and instead choose paragraph-level segmentation, as most datapoints in \benchmarkname{} require reasoning across multiple sentences. We find this change not only improves the overall performance but also accelerates inference speed. Apart from this adjustment, we adhere to the default model and hyperparameters provided by the authors. SummaC-ZS returns a score between -1 and 1.

\paragraph{SummaC-CV} \cite{laban-etal-2022-summac} extends SummaC-ZS by considering all entailment scores for each summary sentence $c_i$. Similar to SummaC-ZS, SummaC-Conv evaluates a summary sentence $c_i$ by dividing the source document $D_i$ into $D_i = \{d_{i,1}, \ldots, d_{i,|d|}\}$. However, instead of selecting the maximum score among all $(d_{i,j}, c_i)$ pairs, SummaC-Conv uses a learned convolutional layer to transform the distribution of entailment scores $\{M(d_{i,j}, c_i):\forall j\}$ into a single score. The final summary score is computed by averaging the scores of individual sentences. As with SummaC-ZS, we use paragraph-level segmentation in our experiments and keep other settings as default. SummaC-Conv outputs a score between 0 and 1.

\paragraph{AlignScore} \cite{zha-etal-2023-alignscore} is an entailment-based model that has been trained on data from a wide range of tasks such as NLI, QA, fact verification, and summarization. It works similarly to SummaC-ZS, with the only difference being that it splits a document $D_i = \{d_{i,1}, \ldots, d_{i,|d|}\}$ into sequential chunks at sentence boundaries. Each chunk contains approximately 350 tokens, determined by white space splitting. In our experiments, we use the default model and hyperparameters as provided by the authors. AlignScore outputs a score between 0 and 1.

\paragraph{T5-NLI-Mixed} \cite{honovich-etal-2022-true} is an entailment-based fact-checker built on T5-XXL. It has been trained on a diverse set of NLI datasets and predicts whether a given claim is supported by a document, outputting ``1'' for supported claims and ``0'' for unsupported ones. The final entailment score is calculated as the probability of the model predicting the token ``1''. To optimize its performance on 2 GPUs from our hardware setup, we select a chunk size of 350 tokens according to the T5 tokenizer. T5-NLI-Mixed outputs a score between 0 and 1.

\paragraph{\textsc{MiniCheck-Rbta, MiniCheck-Dbta}} also split a document into chunks at sentence boundaries, with a chunk size of approximately 400 tokens according to RoBERTA and DeBERTa tokenizers. This results in approximately the same chunk size as in AlignScore, which has a chunk size of 350 tokens using white space splitting. The output scores fall within the range of 0 to 1.

\paragraph{\textsc{FT5-ANLI-L}, MiniCheck\textsc{-FT5}} work the same way as T5-NLI-Mixed, but using only one GPU and setting the chunk size to 500 tokens using white space splitting. The output scores fall within the range of 0 to 1.

\subsubsection{Machine Configuration for Specialized Fact-Chekers} \label{sec:machine-config}

We use two \texttt{NVIDIA RTX A6000} GPUs for T5-NLI-Mixed, given its model size, and one GPU for the remaining models, all on our own hardware. According to Lambda,\footnote{Detailed price specifications are available at \href{https://lambdalabs.com/service/gpu-cloud\#\#pricing}{\texttt{https://lambdalabs.com/service/gpu-cloud\#\#pricing}}.} a single \texttt{NVIDIA RTX A6000} GPU costs \$0.8 per hour and has 48 GB VRAM.

\section{Baseline Synthetic Data Generation Methods} \label{sec:baseline-methods}

We describe the simplified methods in generating C2D and D2C datasets, denoted as \textsc{C2D-Simp} and \text{D2C-Simp}. Performance on models trained on those simplified synthetic datasets can be found in Section~\ref{sec:intrinsic-eval}.

The motivation for these models is to capture the performance of a more basic prompting approach, where we simply ask GPT-4 to generate a data instance in one shot. Comparing the performance of this with C2D/D2C helps validate our more sophisticated prompting strategy.

\paragraph{\textsc{C2D-Simp}} To generate the \textsc{C2D-Simp} dataset, we begin by providing GPT-4 with a claim $c$ and asking it to create a supporting document $D$ that requires multiple sentences together to support the claim (see Table~\ref{tab:support-doc-gen-simplified} for the generation prompt). We then ask GPT-4 to minimally modify $D$ to create a new document $D'$, which can support some atomic facts mentioned in $c$ but not all of them. Inspired by the error type definitions from \citet{tang-etal-2023-understanding}, we provide four different revision types to help GPT-4 generate diverse non-supporting documents, covering various reasons for not supporting the claim (see Table~\ref{tab:non-support-doc-gen-simp} for the prompt). As the generated supporting documents for a given claim tend to be similar despite adjusting the model temperature, we do not generate multiple supporting documents for $c$. Instead, for each claim, we generate one supporting and pair it with one non-supporting document. To enhance the diversity of the training data and maintain a comparable dataset size to our C2D method, we randomly select 3,500 claims from Wikipedia with cited web articles, resulting in the \textsc{C2D-Simp} dataset containing 7K datapoints.

\paragraph{\textsc{D2C-Simp}} We start by directly using the summary sentences generated using the chunk-level summarization step of our D2C method (Section~\ref{sec:d2c}). That is, for each human written document, we have three document chunks $\{D_1, D_2, D_3\}$ and corresponding supporting summary sentences $\{c_1, c_2, c_3\}$ generated by GPT-4. For each $(D_i, c_i, 1)$ tuple, we ask GPT-4 to modify the summary sentence $c_i$ such that the edited summary sentence $c'_{i}$ is no longer supported by the document chunk $D_i$. We leverage the editing method from \citet{laban-etal-2023-summedits}, which is used to construct their SummEdit factual consistency evaluation benchmark. The editing prompt is provided in Table~\ref{tab:non-support-doc-gen-d2c-simp}. We sample 7K datapoints from the generated data to construct \textsc{D2C-Simp}.

\section{Training Details} \label{sec:implement-detail}

We include the training details and hyperparemater details in the section. Unless otherwise specified, we use the default hyperparematers of the backbone models. All models are trained using the standard cross-entropy loss function.

For our baseline models: \textsc{FT5-C2D}, \textsc{FT5-D2C}, \textsc{FT5-ANLI-L}, \textsc{FT5-C2D-S}, and \textsc{FT5-D2C-S}, we fine-tune \texttt{flan-t5-large}\footnote{\href{https://huggingface.co/google/flan-t5-large}{\texttt{huggingface.co/google/flan-t5-large}}} for 2 epochs on prepared data described in Section~\ref{sec:exp-setup} and~\ref{sec:analysis}, using a batch size of 4 and a learning rate of 5e-5.

For MiniCheck-\textsc{Rbta}, MiniCheck-\textsc{Dbta} and MiniCheck-\textsc{FT5}, we begin by fine-tuning the tuned RoBERTa-Large model from AlignScore, \texttt{deberta-v3-large}\footnote{\href{https://huggingface.co/microsoft/deberta-v3-large}{\texttt{huggingface.co/microsoft/deberta-v3-large}}}, and \texttt{flan-t5-large} on their respective training data (Section~\ref{sec:our-models}) for 2 epochs, while excluding 7K D2C synthetic data. We use a batch size of 4 with an accumulation step of 2 and a learning rate of 1e-5 for RoBERTa and 5e-5 for the other two models. We then fine-tune these models on 7K D2C synthetic data for 1 epoch, with a batch size of 4 and learning rate of 1e-5.

We observe that following this training pipeline consistently yields higher performance across all three backbone models compared to training on all data simultaneously. We hypothesize that this improvement stems from the fact that the source documents in the D2C dataset are human-written documents, in contrast to the synthetically generated source documents in the C2D dataset. Fine-tuning on these realistic documents at the end helps the models adapt back to a realistic distribution, preventing them from overfitting to synthetic documents and allowing them to perform well on real documents in the benchmark.

\begin{table}
\small
\begin{tabular}{p{0.75\linewidth}c}
\toprule 
\textbf{Prompt Functionality} & \textbf{Ref.} \\
\midrule
Sentence decomposition & Table~\ref{tab:sent-decomp} \\
Atomic fact expansion (C2D) & Table~\ref{tab:atomic-expan} \\
Document generation (C2D) & Table~\ref{tab:support-doc-gen} \\
Supporting doc. generation (\textsc{C2D-Simp}) & Table~\ref{tab:support-doc-gen-simplified} \\
Non-supporting doc. generation (\textsc{C2D-Simp}) & Table~\ref{tab:non-support-doc-gen-simp} \\
Merging atomic facts (C2D, D2C) & Table~\ref{tab:atomic-fact-merge} \\
Chunk-level summarization (D2C, D2C-Simp) & Table~\ref{tab:chunk-summ} \\
Non-supporting doc. generation (\textsc{D2C-Simp}) & Table~\ref{tab:non-support-doc-gen-d2c-simp} \\
Entailment check for data construction & Table~\ref{tab:entailment-check} \\
\midrule
Zero-shot factual consistency evaluation & Tabel~\ref{tab:zero-shot-eval} \\
Sentence decontextualization & Table~\ref{tab:sent-decontext} \\

\bottomrule
\end{tabular}
\caption{References to prompts. Upper: prompts for our synthetic data generation methods, and simplified synthetic data generation methods. Lower: Prompts for evaluation on \benchmarkname{}.}  \label{tab:prompt-collection}
\end{table}

\section{Prompts} \label{sec:prompts}

In Table~\ref{tab:prompt-collection}, we present the full list of the prompts used throughout our work. We use GPT-3.5 for sentence decomposition and merging atomic facts, and GPT-4 for the remaining prompts. We next elaborate on how we ensure the labeling quality of our synthetically generated data.

\subsection{Quality Assurance for Generations} \label{sec:quality-check}

\paragraph{Sentence decomposition} We adapt a few-shot sentence decomposition prompt from \cite{kamoi-etal-2023-wice}, which can generate complete and correct atomic facts most of the time according to their human evaluation. The prompt (Table~\ref{tab:sent-decomp}) is used for both of our synthetic data generation methods and the claim decomposition experiment in Section~\ref{sec:claim-decmp}.

\paragraph{C2D: Atomic fact expansion} We use a 4-shot prompt (Table~\ref{tab:atomic-expan}) for this step, where we ask GPT-4 to produce a sentence pair where the atomic fact is supported if and only if the information from both sentences is combined. To ensure the quality of the generation, we verify the correctness of this condition after generation via an entailment check by GPT-4 (Table~\ref{tab:entailment-check}). If the correctness is not met, we iterate the process and regenerate a new sentence pair up to a specified number of attempts. In cases where the correctness criterion remains unmet after the specified attempts, we remove the datapoint from the dataset.

\paragraph{C2D: Supporting document generation} We ensure that all sentences $\mathbf{s}$ from the generated sentence pairs are mentioned in (and hence are entailed by) the generated document $D$ by using the entailment check by GPT-4. Same as above, if the document fails to mention all sentences from the sentence pairs, we iteratively generate new documents until a specified number of attempts is reached. It is important to note that we only verify whether sentences $\mathbf{s}$ are mentioned in $D$, as we believe GPT-4 can perform well on this simple task. By construction, if all sentences are mentioned in $D$, then $c$ is supported by $D$. However, directly performing an entailment check on the ($D$, $c$) pair with GPT-4 may introduce many labeling errors, which can negatively impact the performance of the trained models.

\begin{table*}
\small
\renewcommand{\tabcolsep}{1.5mm}
\begin{tabular}{p{0.5\linewidth}p{0.4\linewidth}c}
\toprule 
\textbf{Document} & \textbf{Claim} & \textbf{Label} \\
\midrule
\multirow{3}{*}{\parbox{\linewidth}{More than 5,000 individuals, part of a caravan that crossed into Mexico last month, are now seeking asylum and have established a temporary encampment at the Tijuana Stadium as of today. The Tijuana Stadium, known for hosting sporting events, recently underwent renovations that doubled its seating capacity. Prior to these changes, the stadium had a capacity to accommodate 1,500 spectators.}}
& By this date, over 5,000 members of the caravan were staying at the Tijuana Stadium — a structure with a capacity of 3,000. & S \\
\cmidrule(r){2-3}
& By this date, over 5,000 members of the caravan were staying at the Tijuana Stadium. & S \\
\cmidrule(r){2-3}
& The Tijuana Stadium has a capacity of 3,000. & S \\
\midrule
\multirow{3}{*}{\parbox{\linewidth}{More than 5,000 individuals who are part of a caravan that crossed into Mexico last month have now established a temporary encampment at the Tijuana Stadium, where they are reportedly seeking asylum. The stadium, known for hosting sporting events, could originally accommodate 1,500 spectators before it became the site of the makeshift settlement. As of today, the facility is being repurposed to provide the asylum seekers with temporary shelter as they await the processing of their claims.}}
& By this date, over 5,000 members of the caravan were staying at the Tijuana Stadium — a structure with {\color{red3}{a capacity of 3,000}}. & U \\
\cmidrule(r){2-3}
& By this date, over 5,000 members of the caravan were staying at the Tijuana Stadium. & S \\
\cmidrule(r){2-3}
& The Tijuana Stadium has a {\color{red3}{capacity of 3,000}}. & U \\
\\
\midrule
\multirow{3}{*}{\parbox{\linewidth}{As of today, a group of individuals has established a temporary encampment within the premises of the Tijuana Stadium, according to officials. The stadium, which has recently undergone extensive renovations that included an expansion to double its original capacity, can now welcome a significantly larger audience. Prior to the upgrade, the Tijuana Stadium was known to have a seating capacity for 1,500 spectators; the recent improvements are expected to enhance its utility for various events and gatherings.}}
& By this date, {\color{red3}{over 5,000 members of the caravan}} were staying at the Tijuana Stadium — a structure with a capacity of 3,000. & U \\
\cmidrule(r){2-3}
& By this date, {\color{red3}over 5,000 members of the caravan} were staying at the Tijuana Stadium. & U \\
\cmidrule(r){2-3}
& The Tijuana Stadium has a capacity of 3,000. & S \\
\\
\\
\midrule
\multirow{3}{*}{\parbox{\linewidth}{In a significant movement at the border, a caravan comprising over 5,000 asylum seekers penetrated Mexico's boundaries last month, bringing to the forefront the ongoing challenges faced by migrants from multiple origins. The group has today established a makeshift camp within the confines of the Tijuana Stadium, a venue known for its recent renovation that doubled its seating capacity. The temporal shift marks a new chapter for the individuals on their quest for safety and stability, with the stadium offering a transient sanctuary as they navigate their next steps.}}
& By this date, over 5,000 members of the caravan were staying at the Tijuana Stadium — a structure with {\color{red3}a capacity of 3,000}. & U \\
\cmidrule(r){2-3}
& By this date, over 5,000 members of the caravan were staying at the Tijuana Stadium. & S \\
\cmidrule(r){2-3}
& The Tijuana Stadium has {\color{red3}a capacity of 3,000}. & U \\
\\
\\
\bottomrule
\end{tabular}
\caption{\textbf{Examples using the C2D method.} Documents are generated from the claim (from Wikipedia): \emph{By this date, over 5,000 members of the caravan were staying at the Tijuana Stadium — a structure with a capacity of 3,000.} The same claim can be both supported (S) and unsupported (U) by documents, which encourage models to pay attention to multiple atomic facts in a sentence. Determining the factuality labels of claims requires models to reason over multiple sentences. {\color{red3}Unsupporting facts} are highlighted.} \label{tab:C2D-example}
\end{table*}

\begin{table*}
\small
\renewcommand{\tabcolsep}{1.4mm}
\begin{tabular}{p{0.7\linewidth}p{0.2\linewidth}c}
\toprule 
\textbf{Document} & \textbf{Claim} & \textbf{Label} \\
\midrule
\textbf{(Doc Chunk 1)} With the SAG-AFTRA strike settled, the six-month, multi-guild Hollywood labor disruption has finally ended, but the theatrical damage has only begun to surface. Reviewing the films delayed until next year, a rough estimate suggests that the stoppage cost theaters around \$400 million- \$600 million in gross -- more, when including lost concession revenue. "Barbie," "Oppenheimer," "Sound of Freedom," and "Taylor Swift: The Eras Tours" kept the damage from being worse. Immediately following the SAG-AFTRA settlement, Disney announced wholesale delays in its upcoming release schedule. Their revived plans includes only one Marvel title for 2024 ("Deadpool 3," moved to July 26 from May 3), down from the customary three per year from MCU. Disney was among the first studios to announce delays, with Sony already out Wednesday evening with word the third "Venom" film would move from July to November. Related Stories
The good news for theaters is despite it all, 2023 should still reach the \$9 billion in domestic gross hoped for this year. & 2024 box-office hopes dashed by production delays and major title postponements, costing potentially \$500 million.  & S \\
\midrule
\textbf{(Doc Chunk 2; corresponding chunk)} However, any hopes that 2024 might return to 2019 box-office parity are dashed. Grosses from rescheduled titles will help, but production delays will leave substantial gaps. Even so: It could have been worse. "Dune: Part 2" (Warner Bros.) and "Kraven the Hunter" and "Ghostbusters: Frozen Empire" (Sony) will cost this year's total the most -- perhaps \$400 million ("Ghostbusters" would have had only 12 days of 2023 play). Figure other films, mostly limited/specialized entries like Luca Guadagnino's "Challengers" and Jeff Nichols "The Bikeriders" (Disney), Ethan Coen's "Drive Away Dolls" (Focus), and "The Book of Clarence" (Sony) could have contributed \$100 million or more while riding the awards wave. The biggest unknown is how much the lack of promotion hurt the films released during the strikes. By the time the SAG-AFTRA strike began July 11, most of the summer's top titles had already been released or were about to be, which meant their promotional pushes were all but complete. & 2024 box-office hopes dashed by production delays and major title postponements, costing potentially \$500 million.  & S \\
\midrule
\textbf{(Doc Chunk 2; corresponding chunk)} \sout{However, any hopes that 2024 might return to 2019 box-office parity are dashed.} Grosses from rescheduled titles will help, but production delays will leave substantial gaps. Even so: It could have been worse. "Dune: Part 2" (Warner Bros.) and "Kraven the Hunter" and "Ghostbusters: Frozen Empire" (Sony) will cost this year's total the most -- perhaps \$400 million ("Ghostbusters" would have had only 12 days of 2023 play). Figure other films, mostly limited/specialized entries like Luca Guadagnino's "Challengers" and Jeff Nichols "The Bikeriders" (Disney), Ethan Coen's "Drive Away Dolls" (Focus), and "The Book of Clarence" (Sony) could have contributed \$100 million or more while riding the awards wave. The biggest unknown is how much the lack of promotion hurt the films released during the strikes. By the time the SAG-AFTRA strike began July 11, most of the summer's top titles had already been released or were about to be, which meant their promotional pushes were all but complete. & {\textcolor{red3}{2024 box-office hopes}} dashed by production delays and major title postponements, costing potentially \$500 million.  & U \\
\midrule
\textbf{(Doc Chunk 2; corresponding chunk)} However, any hopes that 2024 might return to 2019 box-office parity are dashed. Grosses from rescheduled titles will help, but production delays will leave substantial gaps. Even so: It could have been worse. "Dune: Part 2" (Warner Bros.) and "Kraven the Hunter" and "Ghostbusters: Frozen Empire" (Sony) will cost this year's total the most -- perhaps \$400 million ("Ghostbusters" would have had only 12 days of 2023 play). \sout{Figure other films, mostly limited/specialized entries like Luca Guadagnino's "Challengers" and Jeff Nichols "The Bikeriders" (Disney), Ethan Coen's "Drive Away Dolls" (Focus), and "The Book of Clarence" (Sony) could have contributed \$100 million or more while riding the awards wave.} The biggest unknown is how much the lack of promotion hurt the films released during the strikes. By the time the SAG-AFTRA strike began July 11, most of the summer's top titles had already been released or were about to be, which meant their promotional pushes were all but complete. & 2024 box-office hopes dashed by production delays and major title postponements, costing potentially {\color{red3}{\$500 million}}.  & U \\
\midrule
\textbf{(Doc Chunk 3)} Some, like DC Comics' "Blue Beetle" (WB) and "The Equalizer 3" (Sony), may have suffered more. Would promotion for all strike-period releases have total \$100 million? Maybe. Theaters were fortunate that both "Barbie" and "Oppenheimer" already had enormous publicity before actors struck, and both had major, Oscar-nominated directors to carry the ball. Their \$950 million combined domestic gross more than doubled expectations. Add the mid-summer sleeper success of "Sound of Freedom" and July and August both were strong months. A wild card, unknown when the strike began, was Taylor Swift's concert film. It certainly filled an October void that existed before the strike and its SAG-AFTRA waver meant she could promote it. The outside-studio success of "Sound of Freedom" and "The Eras Tour" were not only welcome for their grosses, but also because they show it's possible to find releases outside the studios. & \textcolor{red3}{2024 box-office hopes dashed by production delays and major title postponements, costing potentially \$500 million.}  & U \\
\bottomrule
\end{tabular}
\caption{\textbf{Examples using the D2C method.} The full document is from \href{https://www.indiewire.com/news/box-office/strikes-over-damage-theatrical-box-office-1234925398/}{this website}. The same claim (summary sentence) can be supported by both its directly associated document chunk (chunk 2) and a separate chunk (chunk 1) originating from the same document, which has been divided into three distinct chunks. Some \sout{sentences} are removed from the chunk to make the claim unsupported. {\color{red3}Unsupporting facts} are highlighted.} \label{tab:D2C-example}
\end{table*}

\begin{table*}
\small
\begin{tabular}{p{\linewidth}}
\toprule 
Segment the following sentence into individual facts:\\
\\
Sentence: Other title changes included Lord Steven Regal and The Nasty Boys winning the World Television Championship and the World Tag Team Championship respectively.\\
Facts:\\
- Lord Steven Regal won the World Television Championship. \\
- The Nasty Boys won the World Tag Team Championship.\\
\\
Sentence: The parkway was opened in 2001 after just under a year of construction and almost two decades of community requests.\\
Facts:\\
- The parkway was opened in 2001.\\
- The parkway was opened after just under a year of construction.\\
- The parkway was opened after two decades of community requests.\\
\\
Sentence: Touring began in Europe in April-June with guitarist Paul Gilbert as the opening act, followed by Australia and New Zealand in July, Mexico and South America in late July-August, and concluding in North America in October-November.\\
Facts:\\
- Touring began in Europe in April-June.\\
- The opening act of the tour was guitarist Paul Gilbert.\\
- The tour was in Australia and New Zealand in July.\\
- The tour was in Mexico and South America in late July-August.\\
- The tour was concluded in North America in October-November.\\
\\
Sentence: In March 2018, the company partnered With Amazon Web Services (AWS) to offer Al-enabled conversational solutions to customers in India.\\
Facts:\\
- The company partnered with Amazon Web Services (AWS) in March 2018.\\
- The two companies partnered to offer Al-enabled conversational solutions to customers in India.\\
\\
Sentence: The most significant of these is in Germany, which now has a Yazidi community of more than 200,000 living primarily in Hannover, Bielefeld, Celle, Bremen, Bad Oeynhausen, Pforzheim and Oldenburg.\\
Facts:\\
- The most significant of these is in Germany.\\
- Germany now has a Yazidi community of more than 200,000.\\
- Yazidi community in Germany lives primarily in Hannover, Bielefeld, Celle, Bremen, Bad Oeynhausen, Pforzheim and Oldenburg.\\
\\
Sentence: A previous six-time winner of the Nations' Cup, Sebastian Vettel became Champion of Champions for the first time, defeating Tom Kristensen, who made the final for the fourth time, 2-0.\\
Facts:\\
- Sebastian Vettel is a previous six-time winner of the Nations' Cup.\\
- Sebastian Vettel became Champion of Champions for the first time, defeating Tom Kristensen, 2-0.\\
- Tom Kristensen made the final for the fourth time.\\
\\
Sentence: [SENTENCE]\\
Facts:\\
\bottomrule
\end{tabular}
\caption{Sentence decomposition prompt adapted from ~\cite{kamoi-etal-2023-wice}.} \label{tab:sent-decomp}
\end{table*}

\begin{table*}
\small
\begin{tabular}{p{\linewidth}}
\toprule 
Your task is to generate a pair of sentences so that the provided claim can be entailed by the sentence pair. You must make sure that the claim can only be deduced by combining the information from the two sentences that contain unique information. \\
\\
Examples: \\
Provided Claim: The investigation is into allegations that his mayoral campaign received illegal foreign funds.\\
Sentence 1: During the period leading up to the mayoral election, there was a notable increase in his campaign's financial resources.\\
Sentence 2: Investigation shows the funds having origins beyond national boundaries, a detail raising questions under current campaign laws.\\
\\
Provided Claim: Approximately 1,000 fans fainted at the concert.\\
Sentence 1: Emergency services reported an unusually high number of calls for medical assistance during the concert with an attendance of 20,000.\\
Sentence 2: Venue officials estimated that approximately 5\% of the audience required medical attention for fainting.\\
\\
Provided Claim: The interest rate hikes were intended to manage inflation and moderate economic growth.\\
Sentence 1: Central bank officials expressed concern over the rising consumer price index and the overheating of the economy.\\
Sentence 2: The monetary policy committee decided to adjust the interest rates as a response to these economic indicators.\\
\\
Provided Claim: Several advertisers are considering halting their ads on social media platform X.\\
Sentence 1: Some companies are re-evaluating their marketing strategies to avoid association with platforms that fail to address misinformation.\\
Sentence 2: Recent reports show that platform X has received criticism for its handling of false information spreading unchecked.\\
\\
Please make sure that NEITHER sentence alone supports the claim.\\
\\
Your turn:\\
Provided Claim: [CLAIM]\\
\bottomrule
\end{tabular}
\caption{Prompt for \emph{Step 2: Atomic fact expansion} for the C2D method (Section~\ref{sec:c2d}).} \label{tab:atomic-expan}
\end{table*}

\begin{table*}
\small
\begin{tabular}{p{\linewidth}}
\toprule 
We are creating a news article (one paragraph) in the style of The New York Times. We will give you a list of facts to use when writing your article. You must include all the facts in the list. Never state deduced facts or conclusions. The article should stick to the fact list pretty closely. Include as many sentences as needed to write each fact from the list of facts.\\
\\
Facts you must include:\\
-\{FACT1\} \\
-\{FACT2\} \\
... \\
\bottomrule
\end{tabular}
\caption{Prompt for \emph{Step 3: Document generation} for the C2D method (Section~\ref{sec:c2d}).} \label{tab:support-doc-gen}
\end{table*}

\begin{table*}
\small
\begin{tabular}{p{\linewidth}}
\toprule 
Source: [SOURCE] \\
Claim: [CLAIM]\\
\\
Is the claim fully entailed, or implied, by the source? Please answer with "yes" or "no".\\
\bottomrule
\end{tabular}
\caption{Prompt for all steps in our methods that require entailment check.} \label{tab:entailment-check}
\end{table*}

\begin{table*}
\small
\begin{tabular}{p{\linewidth}}
\toprule 
Document: \\
$[$DOCUMENT$]$\\
\\
Please generate a summary for the document with the following requirements:\\
1. The summary should be a fluent and grammatical sentence.\\
2. The summary should be no more than 15 words.\\
3. The summary should cover information across the document.\\
Summary:\\
\bottomrule
\end{tabular}
\caption{Summarization prompt for \emph{Step 1: Chunk-level summarization} for the D2C method (Section~\ref{sec:d2c}) and \textsc{D2C-Simp} method.} \label{tab:chunk-summ}
\end{table*}

\begin{table*}
\small
\begin{tabular}{p{\linewidth}}
\toprule 
Merge the following individual facts into a single sentence:\\
\\
Facts:\\
- Lord Steven Regal wan the World Television Championship. \\
- The Nasty Boys wan and the World Tag Team Championship. \\
Sentence: Other title changes included Lord Steven Regal and The Nasty Boys winning the World Television Championship and the World Tag Team Championship respectively.\\
\\
Facts:\\
- The parkway was opened in 2001.\\
- The parkway was opened after just under a year of construction.\\
- The parkway was opened after two decades of community requests.\\
Sentence: The parkway was opened in 2001 after just under a year of construction and almost two decades of community requests.\\
\\
Facts:\\
- Touring began in Europe in April-June.\\
- The opening act was guitarist Paul Gilbert.\\
- There was a tour in Australia in July.\\
- There was a tour in New Zealand in July.\\
- There was a tour in Mexico in late July-August.\\
- There was a tour in South America in late July-August\\
- The tour was concluded in North America in October-November.\\
Sentence: Touring began in Europe in April-June with guitarist Paul Gilbert as the opening act, followed by Australia and New Zealand in July, Mexico and South America in late July-August, and concluding in North America in October-November.\\
\\
Facts:\\
- The company partnered with Amazon Web Services (AWS) in March 2018.\\
- The two companies partnered to offer Al-enabled conversational solutions to customers in India.\\
Sentence: In March 2018, the company partnered With Amazon Web Services (AWS) to offer Al-enabled conversational solutions to customers in India.\\
\\
Facts:\\
- The most significant of these is in Germany.\\
- Germany now has a Yazidi community of more than 200,000.\\
- Yazidi community in Germany lives primarily in Hannover.\\
- Yazidi community in Germany lives primarily in Bielefeld.\\
- Yazidi community in Germany lives primarily in Celle.\\
- Yazidi community in Germany lives primarily in Bremen.\\
- Yazidi community in Germany lives primarily in Bad Oeynhausen.\\
- Yazidi community in Germany lives primarily in Pforzheim.\\
- Yazidi community in Germany lives primarily in Oldenburg.\\
Sentence: The most significant of these is in Germany, which now has a Yazidi community of more than 200,000 living primarily in Hannover, Bielefeld, Celle, Bremen, Bad Oeynhausen, Pforzheim and Oldenburg.\\
\\
Facts:\\
- Sebastian Vettel is a previous six-time winner of the Nations' Cup.\\
- Sebastian Vettel became Champion of Champions for the first time.\\
- Sebastian Vettel defeated Tom Kristensen.\\
- Tom Kristensen made the final for the fourth time.\\
- The score was 2-0.\\
Sentence: A previous six-time winner of the Nations' Cup, Sebastian Vettel became Champion of Champions for the first time, defeating Tom Kristensen, who made the final for the fourth time, 2-0.\\
\\
Facts:\\
-\{FACT1\} \\
-\{FACT2\} \\
... \\
Sentence:\\
\bottomrule
\end{tabular}
\caption{Merging prompt for \emph{Step 5: Pairing subclaims and generated documents} for the C2D method (Section~\ref{sec:c2d}) and \emph{Step 2: Claim decomposition and subclaim augmentation} for the D2C method (Section~\ref{sec:d2c}).} \label{tab:atomic-fact-merge}
\end{table*}

\begin{table*}
\small
\begin{tabular}{p{\linewidth}}
\toprule 
Determine whether the provided claim is consistent with the corresponding document. Consistency in this context implies that all information presented in the claim is substantiated by the document. If not, it should be considered inconsistent.\\
Document: [DOCUMENT]\\
Claim: [CLAIM]\\
Please assess the claim's consistency with the document by responding with either "yes" or "no".\\
Answer:\\
\bottomrule
\end{tabular}
\caption{Zero-shot factual consistency evaluation prompt for all LLMs.} \label{tab:zero-shot-eval}
\end{table*}

\begin{table*}
\small
\begin{tabular}{p{\linewidth}}
\toprule 
You are provied with a context and a claim. Please first determine if the claim can stand alone whitout the conext. If not, provide a decontextualzied version of the claim that incorporates necessary information from the context to make it self-contained. The revision should be as minimum as possible. Please respond with a JSON format: \{"label": "yes"/"no", "decontext": "NA"/decontextualized claim\}.\\
\\
Example 1: \\
Context: There are many reasons why poetry is important for children. Poetry can help children build confidence through memorizing and reciting poems. It can also provide an easy way for children to remember a lesson or value.\\
Claim: It can also provide an easy way for children to remember a lesson or value.\\
Answer: \{"label": "no", "decontext": "Poetry can provide an easy way for children to remember a lesson or value."\}\\
\\
Example 2:\\
Context: Yes, ancient societies had concepts of rights. The concept of rights first appeared in the theory of natural law which existed in the state of nature. In this state, people enjoyed certain rights sanctioned by natural law.\\
Claim: In this state, people enjoyed certain rights sanctioned by natural law.\\
Answer: \{"label": "no", "decontext": "In the state of nature, people enjoyed certain rights sanctioned by natural law"\}\\
\\
Example 3:\\
Context: The ancient Greeks had some concept of human rights, although there is no single word in classical Greek that captures the sense of "rights" as it is used in modern political thought. However, Greek customs and institutions provided protection to private property unique in the ancient world, instilling a strong sense of equality. The idea of human rights spread quickly from Babylon to Greece and eventually Rome, where the concept of "natural law" arose.\\
Claim: The idea of human rights spread quickly from Babylon to Greece and eventually Rome, where the concept of "natural law" arose.\\
Answer: \{"label": "yes", "decontext": "NA"\}\\
\\
Your Turn:\\
Context: [CONTEXT]\\
Claim: [CLAIM]\\
Answer:\\
\bottomrule
\end{tabular}
\caption{Decontextualization prompt for GPT-4.} \label{tab:sent-decontext}
\end{table*}

\begin{table*}
\small
\begin{tabular}{p{\linewidth}}
\toprule 
We are creating a news article (one paragraph) in the style of The New York Times. We will give you a claim that must be covered when writing your article. All information in the claim must be supported by weaving together various pieces of evidence within the text. That is, the claim should not be directly supported by using one sentence from the article. The generated article should be around 140 words.\\
\\
Claim: [CLAIM]\\
Article:\\
\bottomrule
\end{tabular}
\caption{Supporting document generation prompt for the simplified data generation method \textsc{C2D-Simp}.} \label{tab:support-doc-gen-simplified}
\end{table*}

\begin{table*}
\small
\begin{tabular}{p{\linewidth}}
\toprule 
You are presented with a claim and an article that fully support the claim. You task is to minimally modify the article with the following requirements:\\
\\
1. The modified article no longer fully supports the claim. Some (but not all) statements in the claim should be supported by the modified article.\\
2. The edited article looks close to the original claim.\\
3. The edited claim article should have the similar length with the original article.\\
\\
The followings are the type of revisions you can use to revise the article:\\
- Entity revision: An entity (like a person, place, organization, etc.) from a claim is being edited or not mentioned in the revised article.\\
- Number revision: A number from a claim is being edited or not mentioned in the revised article.\\
- Attribute revision: A syntax unit (either a word, phrase or clause) that modifies a noun is being edited or not mentioned in the revised article.\\
- Predicate revision: A main content verb or content like adverbs that closely relate to the verb is being edited or not mentioned in the revised article.\\
\\
Claim: [CLAIM] \\
Article: [ARTICLE] \\
\\
Please respond in a JSON format: \{``revision\_type'': ..., ``revised\_article'': ...\}.\\
\bottomrule
\end{tabular}
\caption{Nonsupporting document generation prompt for the simplified data generation method \textsc{C2D-Simp}.} \label{tab:non-support-doc-gen-simp}
\end{table*}

\begin{table*}
\small
\begin{tabular}{p{\linewidth}}
\toprule 
Document: \\
$[$DOCUMENT$]$\\
\\
Consistent Summary:\\
$[$CONSISTENT\_SUMMARY$]$\\
\\
Given the document and consistent summary above, generate 10 slightly modified versions of the summary such that the modifications introduce a factual inconsistency. For example, you can modify a number, date, or entity, and negate or modify a statement. Here are some rules to follow:\\
- Each modification should change at most 3-4 words from the original summary, and keep the rest the same.\\
- Each modification should change a different part of the original summary.\\
- Your modifications should be challenging to detect: modify minimally while still introducing a factual inconsistency.\\
- The factual inconsistency you introduce should be subtle. For example, if you replace an entity, make sure you replace it with another entity from the document.\\
- Each modification should start with ``[FIRST\_THREE\_WORDS] [...]'', and end with ``[LAST\_THREE\_WORDS]''\\
\\
Please respond in a JSON format with the following structure:\\
\{``inconsistent\_summaries'': $[$``First inconsistent summary'', ``Second inconsistent summary'', ...$]$\}\\
\bottomrule
\end{tabular}
\caption{Nonsupporting document generation prompt for the simplified data generation method \textsc{D2C-Simp}. The prompt is adapted from SummEdit \cite{laban-etal-2023-summedits}.} \label{tab:non-support-doc-gen-d2c-simp}
\end{table*}

\begin{table*}
\small
\begin{tabular}{p{\linewidth}}
\toprule 
Determine whether each of the provided claims are consistent with the corresponding document. Consistency in this context implies that all information presented in a claim is substantiated by the document. If not, it should be considered inconsistent.\\
Document: [DOCUMENT]\\
Claims: [CLAIM]\\
\\
Claims are displayed with sentence indices. Please evaluate each claim's consistency with the document by responding with either ``yes'' or ``no'' in the JSON format: \{``[1]'': ..., ``[2]'': ..., ...\}.\\
Answer:\\
\bottomrule
\end{tabular}
\caption{Prompt for predicting the factuality labels of all claims in a response for a provided document. This is used mainly for text summarization where multiple summary sentences share the same document.} \label{tab:summ-prompt}
\end{table*}

\end{document}